\pdfoutput=1

\documentclass[11pt]{article}

\usepackage[]{acl}

\usepackage{times}
\usepackage{latexsym}

\usepackage[T1]{fontenc}

\usepackage[utf8]{inputenc}

\usepackage{booktabs}

\usepackage{hyperref}
\usepackage{url}
\usepackage{graphicx}
\usepackage{multicol}
\usepackage{multirow}
\usepackage{subfig}

\usepackage{microtype}

\usepackage{amsmath,amsfonts,bm}

\def\eqref#1{equation~\ref{#1}}

\def\1{\bm{1}}

\DeclareMathAlphabet{\mathsfit}{\encodingdefault}{\sfdefault}{m}{sl}
\SetMathAlphabet{\mathsfit}{bold}{\encodingdefault}{\sfdefault}{bx}{n}

\title{
\textit{Lifelong Pretraining}: Continually Adapting Language Models\\ to Emerging Corpora
}

\author{Xisen Jin$^{\dagger1}$ \quad Dejiao Zhang$^2$ \quad Henghui Zhu$^2$ \quad Wei Xiao$^2$ \\
        \textbf{Shang-Wen Li}$^{\ddagger2}$ \quad \textbf{Xiaokai Wei}$^2$ \quad \textbf{Andrew Arnold}$^2$ \quad \textbf{Xiang Ren}$^1$
        \\
        $^{1}$University of Southern California \quad $^{2}$AWS AI Labs\\
\texttt{\{xisenjin,xiangren\}@usc.edu} \quad \\
\texttt{\{dejiaoz, henghui, weixiaow, shangwenl, 
xiaokaiw, anarnld\}} \\
\texttt{@amazon.com}
}

\newcommand\blfootnote[1]{%
  \begingroup
  \renewcommand\thefootnote{}\footnote{#1}%
  \addtocounter{footnote}{-1}%
  \endgroup
}

\begin{document}

\maketitle
\begin{abstract}

Pretrained language models (PTLMs) are typically learned over a large, static corpus and further fine-tuned for various downstream tasks.
However, when deployed in the real world, a PTLM-based model must deal with data distributions that deviate from what the PTLM was initially trained on. 
In this paper, we study a \textit{lifelong language model pretraining} challenge where a PTLM is continually updated so as to adapt to emerging data. 
Over a domain-incremental research paper stream and a chronologically-ordered tweet stream, we incrementally pretrain a PTLM with different continual learning algorithms, and keep track of the downstream task performance (after fine-tuning). We evaluate PTLM's ability to adapt to new corpora while retaining learned knowledge in earlier corpora.
Our experiments show distillation-based approaches to be most effective in retaining downstream performance in earlier domains. The algorithms also improve knowledge transfer, allowing models to achieve better downstream performance over the latest data, and improve temporal generalization when distribution gaps exist between training and evaluation because of time.
We believe our problem formulation, methods, and analysis will inspire future studies towards continual pretraining of language models.


\end{abstract}
\section{Introduction}
\begin{figure}
    \centering
    \includegraphics[width=\linewidth]{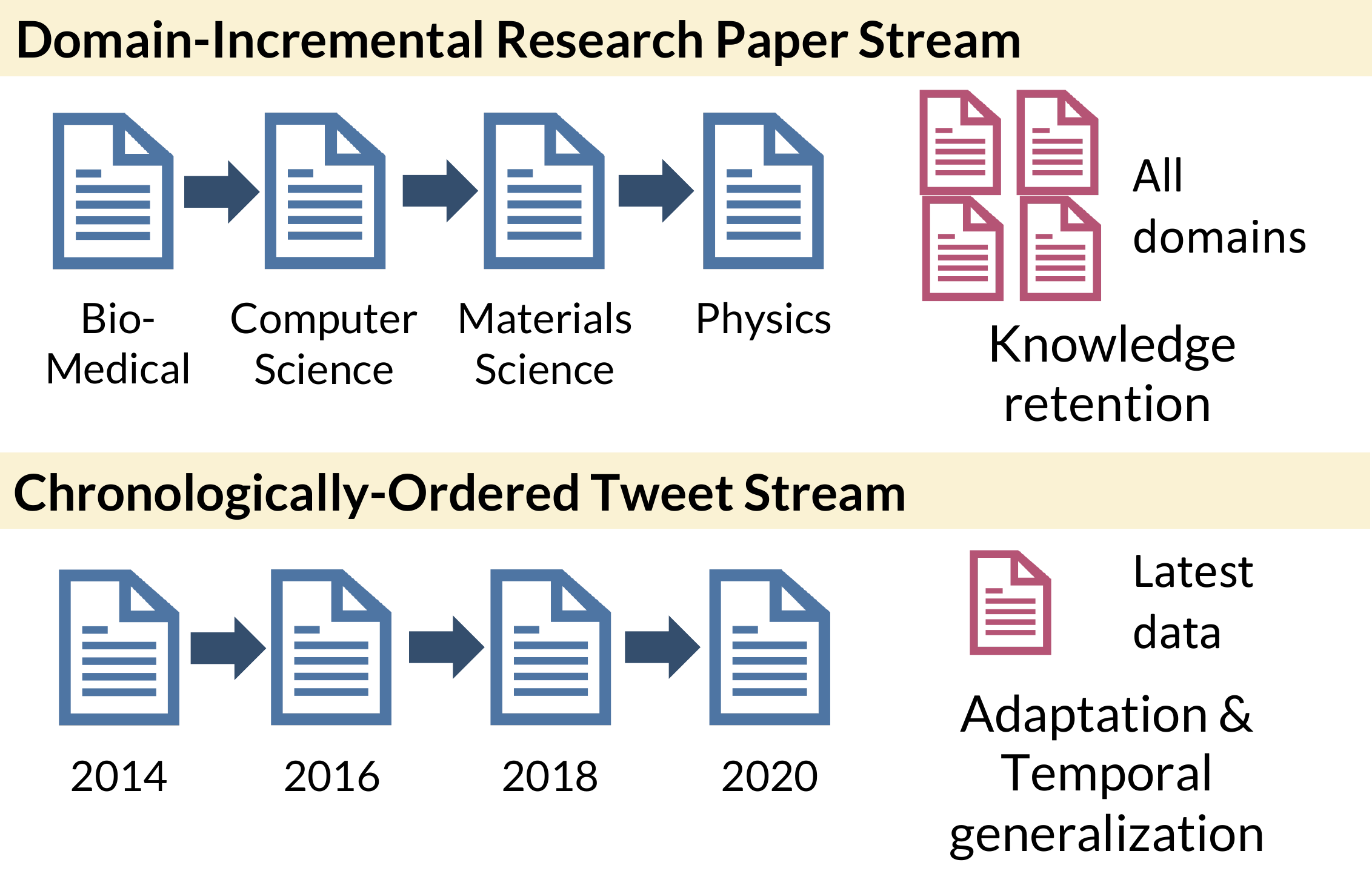}
    \caption{Two data streams created for studying lifelong language model pre-training. We focus on evaluating knowledge retention on the domain-incremental research papers stream; we focus on adaptation to the latest data and temporal generalization on the chronologically ordered tweet stream.}
    \label{fig:datasets}
\end{figure}

\blfootnote{$\dagger$ Work done during an internship at AWS AI Labs.}
\blfootnote{$\ddagger$ Work done while at Amazon.}
Pretrained language models (PTLMs) have achieved remarkable performance on benchmark datasets for a range of NLP tasks~\cite{Liu2019RoBERTaAR, brown2020language}. 
However, when deployed in the wild, NLP systems must deal with emerging data that have constantly shifting data distribution, different from the text corpora they were initially pretrained on --- for example, when new data domains are introduced (upper part of Fig.~\ref{fig:datasets})~\cite{Gururangan2020DontSP}, or when the language uses and vocabulary change over time (lower part of Fig.~\ref{fig:datasets})~\cite{Lazaridou2021PitfallsOS}. 
Fine-tuning from a static and possibly ``outdated" PTLM may limit the model performance on downstream tasks, 
as the PTLM may no longer provide an effective model initialization~\cite{beltagy2019scibert, Mller2020COVIDTwitterBERTAN}.
Here we look to understand whether continuously adapting a PTLM to emerging data can yield gains on various downstream tasks, and how to achieve better downstream performance for such lifelong PTLM adaptation.

A number of recent works make attempts on adapting PTLMs to a new data domain.
\citet{Gururangan2020DontSP, Yao2021AdaptandDistillDS} adapt language models to corpora of different genres and topics
and observe performance improvement in domain-specific downstream tasks.~\citet{Arumae2020AnEI} further show that by regularizing the parameters of PTLMs, the downstream tasks performance on the general domain can be preserved. Another line of works focuses on temporal domain shift~\cite{Hombaiah2021DynamicLM}, which analyzes the effect of pretraining over up-to-date data to the downstream tasks.~\citet{Rttger2021TemporalAO} further study vocabulary composition approaches for improving adaptation to up-to-date corpora.
However, these work focus their study on adapting PTLM to a single new domain; while in practice, corpora from distinct domains and time stamps may emerge sequentially. 
Whether one can maintain a single, up-to-date PTLM remains an open problem. 
Related to this,~\citet{Lazaridou2021PitfallsOS} study adaptation of PTLMs over temporal data streams, but solely focus on language modeling instead of fine-tuning performance. 
It is also important to understand multiple aspects of the utility of lifelong PTLM pretraining, such as knowledge retention over all the seen data, and study what methods can improve the utility of PTLMs in such a continual pretraining process.

In this paper, we formulate a \textit{Lifelong Language Model Pretraining} task to simulate practical scenarios of maintaining and adapting a PTLM over emerging corpora, create a testbed (along with pretraining data streams and downstream tasks) for studying continual pretraining algorithms, and present a systematic evaluation protocol for measuring the progress made on this challenging problem (see Figure~\ref{fig:problem} for an illustration). 
We consider two types of text corpus sequences when constructing pretraining data streams, each of which simulates a representative use case and that has slightly different focuses on the evaluation: continuously learning a single model that is applicable to both old and new domains; and improving the model's ability to handle latest data.
Specifically, we construct
1) a domain-incremental text stream that consists of academic papers published in four research fields, and
2) a temporal tweet stream that consists of tweets collected from four different years.
By conducting systematic experiments on these two data streams,
we look to answer a series of analysis questions: 1) whether continual pretraining retains fine-tuning performance over earlier corpora compared to traditional offline pretraining, 2) whether pretraining improves downstream performance on the latest data, and 3) whether pretraining improves temporal generalization where training and evaluation have distribution gaps because of time.
To address the research questions above, we conduct a systematic evaluation of existing continual learning (CL) algorithms, spanning over model-expansion based, memory-based, and distillation-based approaches. 
Our results show distillation-based approaches are most effective in knowledge retention in the research paper stream, while simultaneously improve adaptation to latest data and temporal generalization in the tweet stream.
We believe our problem formulation, evaluation setup, methods and analysis can inspire more future work on continual pretraining of language models.

\section{Problem Formulation}
\label{sec:setup}Here we present the problem formulation for lifelong pretraining of PTLM, provide details about the data stream construction process and downstream tasks, and introduce the evaluation protocol.


\subsection{Lifelong Pretraining of PTLMs}
We consider the scenario where one needs to deploy and/or maintain NLP models over a sequence of $T$ data domains. At each time step $t$ the model visits an unlabeled text corpus $D_t$ from a domain with a data distribution $P(D_t)$. The data distribution $P(D_t)$ evolves as the time step $t$, forming a \textit{data stream} $D_{1..T}=\{D_1, D_2,...D_T\}$. In practice, the data domain shift can refer to the topic change of the text content (from computer science research papers to biomedical papers), or temporal evolution of the text (from past to recent tweets).
The task of \textit{lifelong pretraining of PTLM} looks to continuously adapt a language model $f$ as the model visits (unlabeled) text corpus $D_t$ from the data stream $D_{1..T}$, in order to provide a good model initialization for fine-tuning on downstream tasks from the same domain. With slight abuse in notations, we also use $D_t$ to directly refer to a data domain. 

Here, we assume a language model $f$ is updated sequentially over each pretraining corpora $D_t$, without accessing the full earlier corpora $\{D_i\}_{i<t}$ in the data stream $D_{1..T}$. This aims to capture practical constraints such as privacy restriction for storing earlier data, or computation budget for training over all the text corpora in $D_{1..T}$. 
We use $f_t$ to denote the language model right \textit{after} updating on the domain $D_t$. 
In our study, $f$ is a RoBERTa-base transformer~\cite{Liu2019RoBERTaAR} and the model ($f_0$) is initialized with pretrained RoBERTa weights.

The utility of the PTLMs $\{f_t\}$ is evaluated based on their fine-tuned model performance on various downstream tasks. 
After updating on a domain $D_{i}$,
the model $f_{i}$ can be fine-tuned over downstream tasks from visited domains $D_t$ where $t \leq i$. We note the set of downstream tasks related to domain $D_t$ as $S_t = \{S_t^j\}_{j=1}^{N_t}$, assuming the number of downstream tasks is $N_t$. 
Note that in the fine-tuning stage, model $f_t$ has no access to any of the pretraining corpus $D_{1..T}$.

\begin{figure}
    \centering
    \includegraphics[width=\linewidth]{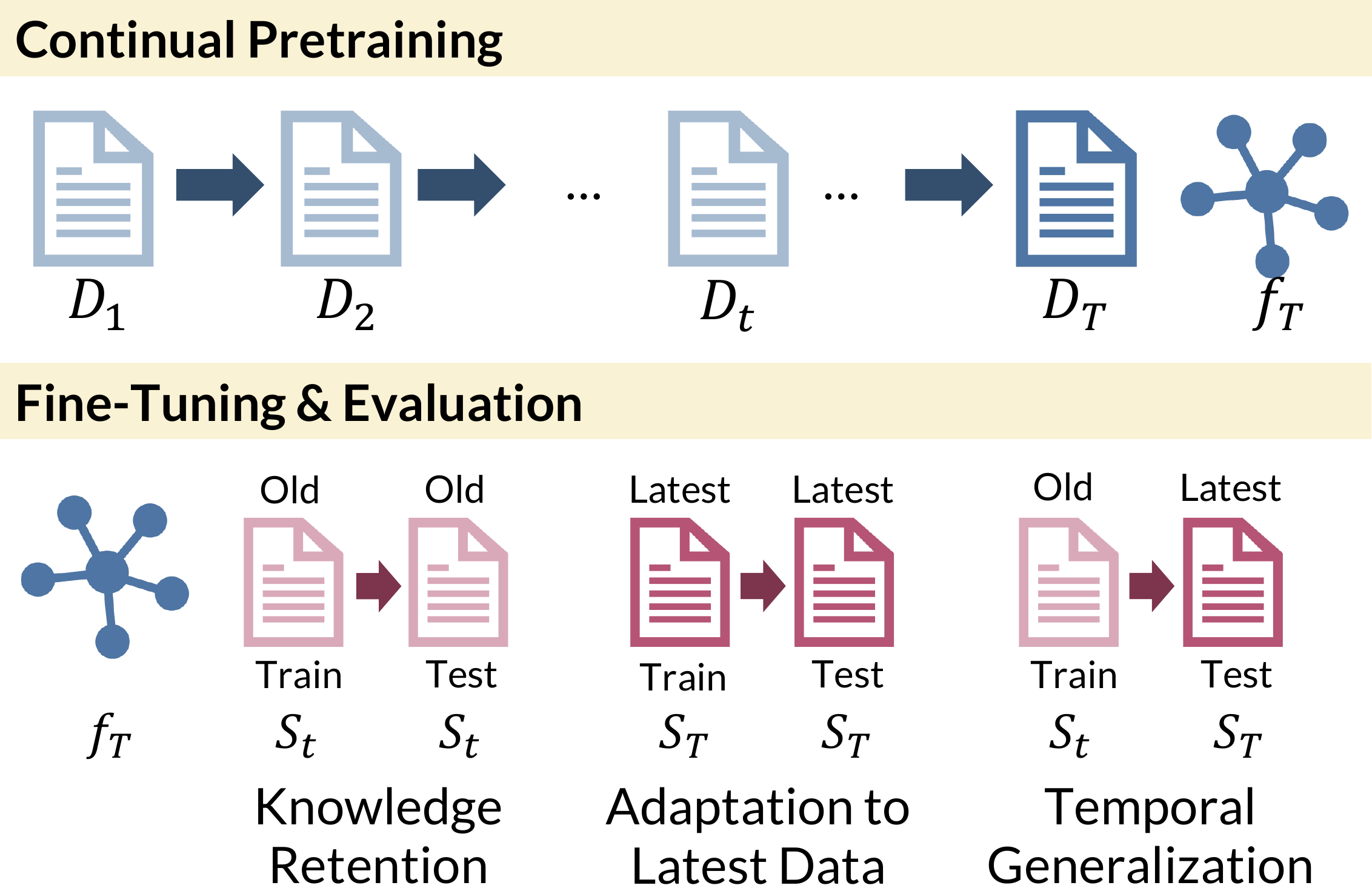}
    \caption{\small \textbf{Training, evaluation setups, and metrics of lifelong language model pretraining.} The model sequentially visits each corpus, and is fine-tuned on downstream datasets related to the domains of pretraining. We evaluate knowledge retention and adaptation to new data with downstream fine-tuning performance on old and latest domains respectively. Besides, we evaluate temporal generalization where training/test examples are drawn from different time steps.
    }
    \label{fig:problem}
\end{figure}

\vspace{-0.1cm}
\subsection{Data Streams \& Downstream Datasets}
\label{ssec:datasets}
We construct data streams to simulate two representative scenarios of data domain shifts in practice (also see Fig.~\ref{fig:datasets}): one \textit{domain-incremental} stream to simulate the sequential changes of research paper areas; and one \textit{chronologically-ordered} stream to simulate tweets emerging over time.

\paragraph{Domain-incremental Paper Stream.} This paper stream consists of the full text of research papers published in four research areas: biomedical, computer science, material science, and physics, filtered from the S2ORC dataset\footnote{\scriptsize{We use the 20200705v1 version of the S2ORC dataset at \url{https://github.com/allenai/s2orc}}}, which are presented sequentially to the model. For each domain, we evaluate downstream performance over two datasets. The downstream tasks span over various tasks such as relation extraction and named entity recognition, and are summarized in Table~\ref{tab:academics_downstream}. We detail these datasets in Appendix~\ref{apdx:dataset_details}. 

\paragraph{Chronologically-ordered Tweet Stream.} 
This tweet data stream consists of tweets from the year 2014, 2016, 2018 and 2020, collected by the Archive Team\footnote{\scriptsize{\url{https://archive.org/details/twitterstream}}} and preprocessed following~\citet{nguyen2020bertweet}. These four tweet corpora are presented sequentially to the language model following the chronological order of the tweet year.
For downstream tasks, we hold out 1M tweets from each year's corpus to construct multi-label hashtag prediction datasets~\cite{Gong2016HashtagRU} and single-label emoji prediction datasets~\cite{Barbieri2018SemEval2T}. On two datasets, we report label ranking average precision scores (a multi-label version of MRR) of models ~\cite{Azeemi2021COVID19TA} and Macro-F1 respectively. The detailed dataset construction process is included in Appendix~\ref{apdx:dataset_details}. 

\subsection{Evaluation Protocol}
\label{ssec:evaluation_protocols}

We consider three key aspects for evaluating the utility of the language models $\{f_t\}$ that are continuously updated over the data stream $D_{1..T}$, also illustrated in Figure~\ref{fig:problem}: 
1) knowledge retention and transfer over the pretraining corpora seen earlier; 2) adaptation to the latest data domain, and 3) temporal generalization when training and evaluation data are from different time steps.


\paragraph{Knowledge Retention.} 
A key utility of continual language model pretraining is to obtain a single model applicable to all domains. We focus on the evaluation of the ability with the domain-incremental paper stream, because for the tweet stream, the practical need of performance over outdated data is limited. Knowledge retention is measured with the downstream task performance from earlier or the current domains that the pretrained model has visited. More formally, for each pretrained model checkpoint in $\{f_i\}$, we fine-tune $f_i$ over downstream tasks $\{S_t\}$ where $t\leq i$ and evaluate the corresponding test set performance. It is important that the models do not suffer from catastrophic forgetting~\cite{Robins1995CatastrophicFR}, \textit{i.e.,} significantly reduced helpfulness when $f_i$ is fine-tuned for downstream tasks $S_t$ from earlier domains with $t<i$. 

\begin{table}[]
\centering
\scalebox{0.65}{
\begin{tabular}{@{}llc@{}}
\toprule
Domains           & Downstream Datasets                       & Metrics  \\ \midrule
Bio-Medicine      & Chemprot~\cite{Vindahl2016ChemProt30AG}   & Micro-F1 \\
                  & RCT-Sample~\cite{Dernoncourt2017PubMed2R} & Micro-F1 \\
Comp. Science  & ACL-ARC~\cite{Jurgens2018MeasuringTE}     & Macro-F1 \\
                  & SciERC~\cite{Luan2018MultiTaskIO}         & Macro-F1 \\
Mat. Science & Synthesis~\cite{mysore2019materials}      & Macro-F1 \\
                  & MNER~\cite{olivetti2020data}              & Micro-F1 \\
Physics           & Keyphrase~\cite{augenstein2017semeval}    & Macro-F1 \\
                  & Hyponym~\cite{augenstein2017semeval}      & Macro-F1 \\ \bottomrule
\end{tabular}
}
\caption{Summary of downstream datasets relevant to each domain in the research paper stream.}
\label{tab:academics_downstream}
\vspace{-0.2cm}
\end{table}

\paragraph{Adaption to Latest Data Domain.} 
In certain scenarios, performance of downstream models over the latest data domain should be emphasized. For example, classifiers in the tweet domain are usually trained and evaluated with up-to-date data for practical deployment. Formally, we focus on the downstream task performance of models fine-tuned from the final pretrained model checkpoint $f_T$, where the downstream tasks $S_T$ are also from the latest domain. To succeed in these metrics, it is crucial for the model to transfer knowledge from earlier domains to the latest domain. 


\paragraph{Temporal Generalization Ability.}
We consider another practical fine-tuning scenario in the tweet stream where the model is trained on outdated data and evaluated on the latest data~\cite{Rijhwani2020TemporallyInformedAO, Huang2018ExaminingTI}, referred to as the \textit{temporal generalization} ability. 
Formally, we fine-tune the final pretrained model checkpoint $f_T$ over the training set of downstream tasks $S_t$ from an earlier time step ($t < T$), and evaluate on the test set of the downstream tasks $S_T$ from the latest time step $T$.



\section{Methods}
\label{sec:methods}

\begin{figure}
    \centering
    \includegraphics[width=\linewidth]{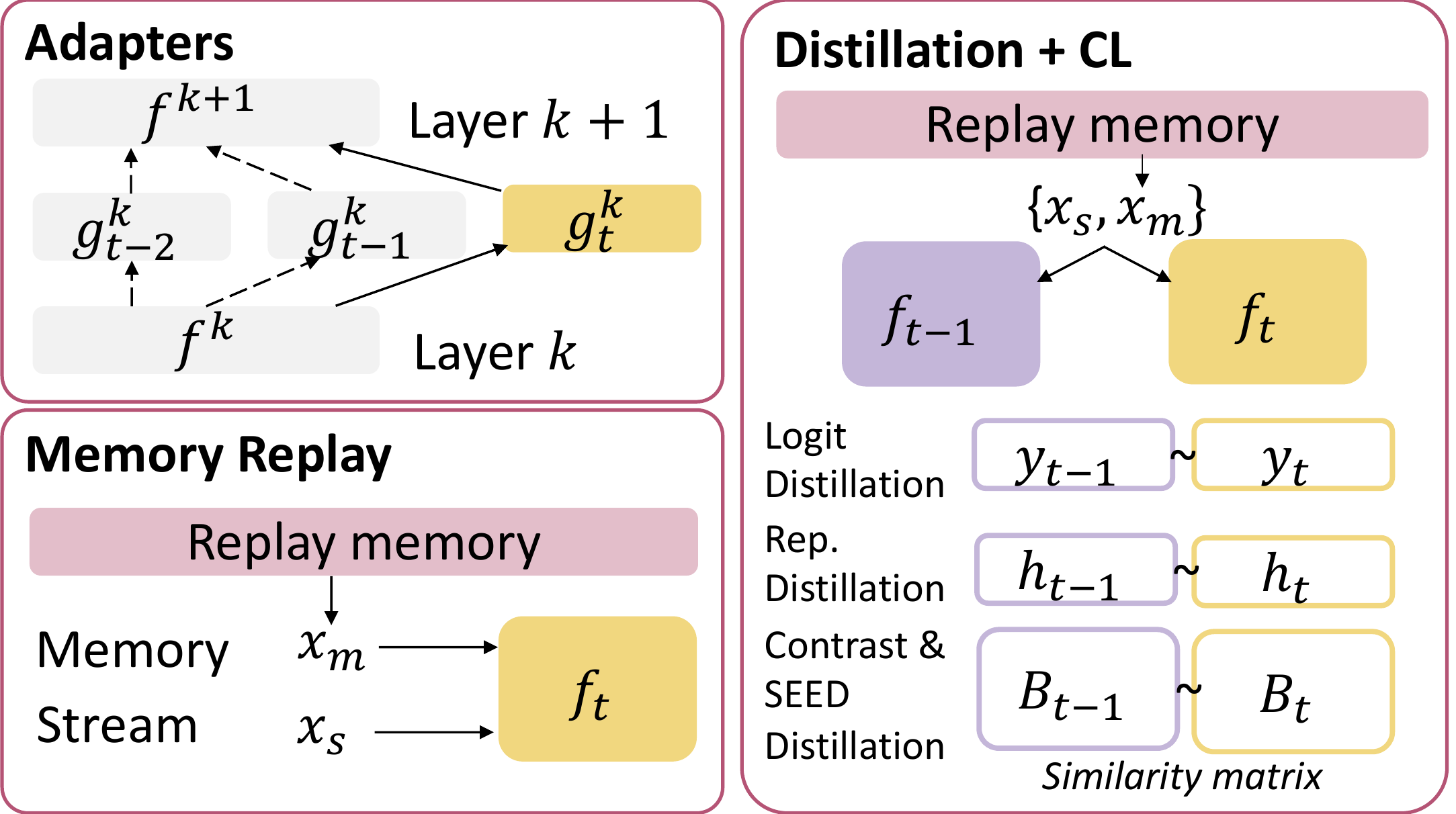}
    \caption{\textbf{Comparison of adapter, memory replay, and distillation-based continual learning algorithms.} 
    Details of the methods are introduced in Sec.~\ref{sec:methods}.
    }
    \label{fig:method}
\end{figure}

Lifelong language model pretraining introduces novel challenges because of the large training sets and more comprehensive evaluation protocols compared to classification tasks. 
We establish several strong baselines, and evaluate the performance of continual learning algorithms from different categories spanning over model-expansion, memory-based, and distillation-based approaches, We illustrate the approaches in Figure~\ref{fig:method}. 

\vspace{-0.1cm}
\subsection{Simple Baselines}
\vspace{-0.1cm}
We consider several simple baselines which continual learning algorithms will be compared against. \texttt{RoBERTa-base} ($f_0$) corresponds to not pretraining on any of the domain-specific corpora. By separately pretraining $f_0$ on each corpus $D_1,D_2,...D_T$, we obtain $T$ \texttt{Task-Specific} pretrained models. 
We also pretrain $f_0$ sequentially over $D_{1..T}$, which we refer to as \texttt{sequential pretraining}. While it allows knowledge transfer between domains compared to domain-specific models, without any continual learning algorithms, sequential pretraining is prone to catastrophic forgetting~\cite{Robins1995CatastrophicFR}. Finally, we randomly shuffle corpora from all domains $D_{1..T}$ before pretraining, noted as \texttt{Multi-Task Learning (MTL)}. MTL corresponds to an offline training paradigm that models new corpora by re-training over all corpora seen before. The drawback is that it requires storing full data from earlier domains, and that it can be extremely costly to repetitively retrain over earlier data if new data keeps emerging.


\vspace{-0.1cm}
\subsection{Model-expansion and Regularization-based Methods}
\vspace{-0.1cm}

We first introduce model-expansion based approaches, which add small trainable modules (\textit{e.g.,} multi-layer perceptron) to the model per new domain while keeping other parts of the model frozen.
The \texttt{Adapter} approach is a representative approach that learns a set of ``adapter'' layers $g_t = \{g_t^k\}_{k=1}^K$ for each domain $D_t$ and each of the $K$ transformer layers~\cite{houlsby2019parameter}. 
We also experiment with a simple \texttt{Layer Expansion} approach, which learns separate top two layers of the transformer and the prediction head for each domain. 
We also involve a regularization-based continual learning baseline, online EWC~\cite{Schwarz2018ProgressC}, which directly penalize change of model parameters. 


\vspace{-0.1cm}
\subsection{Memory Replay Methods}
\label{ssec:mem_replay_methods}
\vspace{-0.1cm}
We also experiment with Experience Replay (ER)~\cite{chaudhry2019tiny}, which alleviates forgetting by storing a subset of earlier examples and periodically re-training (replaying) over them. We maintain a fixed-size memory $M$ ($100k$ examples by default) and populate the memory $M$ each time pretraining on a domain $D_t$ finishes with examples in the current domain. We ensure $M$ always contains a balanced sample of examples from all seen domains $D_{1..{t}}$. We replay a mini-batch of examples from the memory every 10 training steps.

\subsection{Distillation-based CL Methods}
While knowledge distillation (KD)~\cite{Hinton2015DistillingTK} techniques have been studied intensively for pretrained language models~\cite{Sun2019PatientKD}, applying them to continual learning has been under-explored outside image classification tasks~\cite{Li2018LearningWF, Rebuffi2017iCaRLIC, Hou2018LifelongLV}. Distillation based CL approaches store one previous model checkpoint of the model (noted as $f_{t-1}$) and regularize the differences between $f_{t-1}$ and the current model $f_t$.
We adapt several existing knowledge distillation techniques to PTLMs and utilize them for continual learning. We note, while individual distillation techniques are not original, their adaptation to CL algorithms can be novel.

We perform distillation with examples from the current domain $D_{t}$ and a replay memory $M$ (similar to ER). 
Despite the potential gap between $D_{t}$ and the training data of $f_{t-1}$, the approach allows utilizing more data for distillation. Formally,  
each time the model receives a mini-batch of stream examples $\bm{x}_s$ or a draws mini-batch of memory examples $\bm{x}_m$ from $M$ (both noted as $\bm{x}$), we collect certain outputs of the model (\textit{e.g.}, output logits or intermediate representations) with $f_{t-1}$ and $f_t$. 
We compute a distillation loss $\ell_{\textrm{KD}}(\bm{x}, f_{t-1}, f_t)$ that penalizes the differences between the model outputs, and jointly optimize it with the masked language modeling loss $\ell_{\textrm{MLM}}$. The final objective is written as $\ell = \ell_{\textrm{MLM}} + \alpha\ell_{\textrm{KD}}$, where $\alpha$ is a hyperparameter to weight the distillation loss. 


\paragraph{Logit Distillation.} In logit distillation~\cite{Hinton2015DistillingTK}, we collect the output logits of $f_t$ and $f_{t-1}$, noted as $\bm{y}_t$ and $\bm{y}_{t-1}$ respectively. The distillation loss is computed as 
$D_{\textrm{KL}}(\bm{y}_t, \bm{y}_{t-1})$, where $D_{\textrm{KL}}$ is the Kullback–Leibler divergence function.

\paragraph{Representation Distillation.} 
We also consider minimizing the representational deviation of sentences between previous and current models~\cite{Sun2019PatientKD, Jiao2020TinyBERTDB}. We extract the representation of each word of two models, noted as $\bm{h}_{t-1}^{1:N}$ and $\bm{h}_{t}^{1:N}$, before the masked language modeling prediction head, where $N$ is the length of the sentence. Then, we compute MSE loss $\vert\vert \bm{h}_{t-1}^{1:N} -  \bm{h}_{t}^{1:N} \vert\vert_2^2$ as the distillation loss.

\begin{table*}[]
\centering
\scalebox{0.60}{
\begin{tabular}{@{}lcccccccccccc@{}}
\toprule
Task          & \multicolumn{3}{c}{$D_1$ - Biomedical} & \multicolumn{3}{c}{$D_2$ - Computer Science} & \multicolumn{3}{c}{$D_3$ - Materials Science} & \multicolumn{3}{c}{$D_4$ - Physics} \\
\cmidrule(r){1-1} \cmidrule(lr){2-4} \cmidrule(lr){5-7} \cmidrule(lr){8-10} \cmidrule(l){11-13} 
Dataset       & Chemprot           & RCT-Sample    &  MLM      & ACL-ARC               & SciERC        & MLM        & MNER                   & Synthesis    & MLM         & Keyphrase         & Hyponym     & MLM     \\ \midrule
Roberta-base  & 82.03$_{\pm 0.7}$          & 78.07$_{\pm 0.7}$    & 1.993      & 64.32$_{\pm 2.8}$             & 79.07$_{\pm 1.6}$     & 2.153        & 83.15$_{\pm 0.3}$              & 91.25$_{\pm 0.6}$        & 2.117     & 66.21$_{\pm 1.0}$         & 67.59$_{\pm 4.5}$    &   2.278  \\
Sequential Pretraining         & 82.09$_{\pm 0.5}$          & 79.60$_{\pm 0.5}$      & 1.654    & 72.73$_{\pm 2.9}$             & 81.43$_{\pm 0.8}$    & 1.807         & \textbf{83.99$_{\pm 0.3}$}              & 92.10$_{\pm 1.0}$    & 1.590          & \textbf{67.57$_{\pm 1.0}$}         & \textbf{74.68$_{\pm 4.4}$}    & \textbf{1.381}    \\
\midrule
ER            & 82.73$_{\pm 0.3}$          & 79.98$_{\pm 0.3}$  & 1.737        & 72.50$_{\pm 1.0}$             & 81.64$_{\pm 1.1}$   & 1.857           &  \textbf{83.99$_{\pm 0.4}$}              & 92.65$_{\pm 0.4}$     & 1.621        & 66.11$_{\pm 1.1}$         & 72.82$_{\pm 4.3}$   & 1.391      \\
Online EWC          & 81.83$_{\pm 0.2}$          & 78.84$_{\pm 0.5}$  & 1.655        & 71.81$_{\pm 2.6}$             & 80.79$_{\pm 0.5}$   & 1.803           &  83.43$_{\pm 0.4}$              & 91.89$_{\pm 0.5}$     & 1.571        & 66.70$_{\pm 0.6}$         & 72.98$_{\pm 6.0}$   & 1.388      \\
Adapter       & 83.30$_{\pm 0.4}$          & 80.41$_{\pm 0.4}$  & 1.417        & 69.32$_{\pm 3.5}$             & 80.22$_{\pm 1.5}$      & \textbf{1.633}       &          83.91$_{\pm 0.3}$         &    91.69$_{\pm 0.6}$    & 1.522           & 66.23$_{\pm 1.4}$         & 69.65$_{\pm 4.5}$   & 1.554     \\
Layer Expansion & 83.74$_{\pm 0.3}$ & 81.10$_{\pm 0.5}$ & \textbf{1.210} & 65.17$_{\pm 2.9}$ & 79.35$_{\pm 0.8}$  &  1.756 & 82.48$_{\pm 0.4}$ & 92.33$_{\pm 1.0}$ & 1.389 &  65.70$_{\pm 1.1}$ &  73.34$_{\pm 3.7}$ & 1.534 \\
Logit-KD      & 83.39$_{\pm 0.4}$          & \textbf{81.21$_{\pm 0.1}$}  & 1.392       & \textbf{73.70$_{\pm 3.4}$}             & 81.92$_{\pm 0.8}$      & 1.699       & 83.96$_{\pm 0.3}$              & 92.20$_{\pm 1.0}$      & \textbf{1.425}       & 64.75$_{\pm 1.1}$         & 71.29$_{\pm 3.6}$   & 1.460      \\
Rep-KD        & 82.34$_{\pm 0.3}$          & 79.59$_{\pm 0.5}$ & 1.684        & 71.17$_{\pm 2.5}$             & 78.78$_{\pm 1.1}$      & 1.810       & 84.13$_{\pm 0.3}$              & 92.02$_{\pm 0.8}$   & 1.585           &      65.96$_{\pm 1.6}$      & 73.93$_{\pm 5.5}$       &      1.389     \\
Contrast-KD   & 82.29$_{\pm 0.5}$          & 79.92$_{\pm 0.4}$  & 1.722        & 71.15$_{\pm 1.1}$             & 80.49$_{\pm 1.6}$    & 1.856         & 83.26$_{\pm 0.4}$              & 92.62$_{\pm 0.7}$   & 1.612          & 65.95$_{\pm 1.7}$         & 72.26$_{\pm 3.1}$   & 1.428     \\
SEED-KD & 82.78$_{\pm 0.3}$ & 80.38$_{\pm 0.4}$ & 1.720 & 69.98$_{\pm 2.4}$  & 81.61$_{\pm 0.7}$ & 1.829 & 82.99$_{\pm 0.4}$ & 92.35$_{\pm 0.7}$ & 1.609 & 65.35$_{\pm 1.0}$ & 74.79$_{\pm 4.1}$ & 1.401 \\ 
SEED-Logit-KD &  \textbf{83.72$_{\pm 0.4}$} & 81.05$_{\pm 0.2}$  & 1.391        & 69.90$_{\pm 4.5}$             & \textbf{83.03$_{\pm 0.6}$}   & 1.703           & 83.28$_{\pm 0.5}$              & \textbf{92.87$_{\pm 1.0}$}     & 1.428        & 65.96$_{\pm 1.5}$         & 71.92$_{\pm 5.5}$    & 1.460    \\ \midrule
Task-Specific LM        & 83.74$_{\pm 0.3}$          & 81.10$_{\pm 0.5}$    & 1.210      & 72.20$_{\pm 2.6}$             & 81.24$_{\pm 1.7}$   & 1.629          & 84.02$_{\pm 0.2}$              & 91.56$_{\pm 0.4}$        &  1.418     & 65.95$_{\pm 1.1}$         & 69.43$_{\pm 4.5}$  & 1.426      \\
MTL           & 82.91$_{\pm 1.6}$          & 80.67$_{\pm 0.4}$   & 1.289        & 69.46$_{\pm 1.8}$             & 81.12$_{\pm 0.8}$  & 1.616        & 83.92$_{\pm 0.3}$              & 92.66$_{\pm 0.6}$     &  1.355        & 65.37$_{\pm 1.6}$         & 73.31$_{\pm 5.2}$  & 1.418 \\  \bottomrule
\end{tabular}
}
\caption{\small \textbf{Results on the Research Paper stream.} We report log perplexity of MLM and the performance of downstream models fine-tuned from the final checkpoint of the pretrained model ($t=4$). Performance of the best performing CL algorithm is marked bold.
}
\label{tab:academics}
\vspace{-0.2cm}
\end{table*}

\paragraph{Contrastive Distillation.}
In addition to output logits and hidden representations, we further look into \textit{representational similarity within a batch of examples} as additional knowledge to distill. The approach is adapted from~\cite{Cha2021Co2LCC}, which is originally studied for supervised image classification tasks. 
We briefly introduce the adapted algorithm and leave the details in Appendix~\ref{apdx:cl}.
During continual pretraining, in addition to the language model pretraining objective, we add an unsupervised contrastive learning objective, namely the SimCSE~\cite{Gao2021SimCSESC} objective
to encourage sentence representations to reflect semantic similarities between sentences. Then, we compute the intra-batch representational similarity matrices of sentence representations (\textit{i.e.} between each pair of examples in the mini-batch) with $f_{t-1}$ and $f_t$, noted as $\mathbf{B}^{t-1}$ and $\mathbf{B}^{t}$, and minimize the cross entropy loss $\ell_{\text{distill}} = - \frac{1}{N} \sum_{i=1}^{N} \sum_{j=1}^{N} \mathbf{B}_{ij}^{t-1}\log \mathbf{B}_{ij}^{t}$



\paragraph{Self-Supervised Distillation (SEED).} SEED distillation proposed by~\cite{Fang2021SEEDSD} has a similar spirit as the contrastive distillation. The only difference is that it distills representational similarity \textit{between the batch and a large set of other examples}. We leave the details of the algorithm in Appendix~\ref{apdx:cl}. 
We further combine SEED Distillation
with logit distillation and refer to the approach as SEED-Logit Distillation.

\section{Results}
We summarize our findings over the created data streams. We ask whether lifelong pretraining and continual learning algorthms are effective base on our evaluation protocol proposed in Sec.~\ref{ssec:evaluation_protocols}. 

\subsection{Experiment Settings}
We use the \texttt{RoBERTa-base} model~\cite{Liu2019RoBERTaAR}, initialized with RoBERTa-base weights throughout the experiments. We set the maximal sequence length to 128 and an effective training batch size of 2,048. On the research paper stream, models are trained for 8$k$ steps in the first domain and 4$k$ steps in the subsequent domains. On the Tweet stream, we train the models for 4$k$ steps in each domain. These correspond to less than a single pass of data in each domain. See Appendix~\ref{sec:hyps} for detailed setups. 

\subsection{Domain Incremental Data Stream}
\begin{figure*}[!t]
    \centering
    \subfloat[\scriptsize{Chemprot}]{\includegraphics[width=0.48\linewidth]{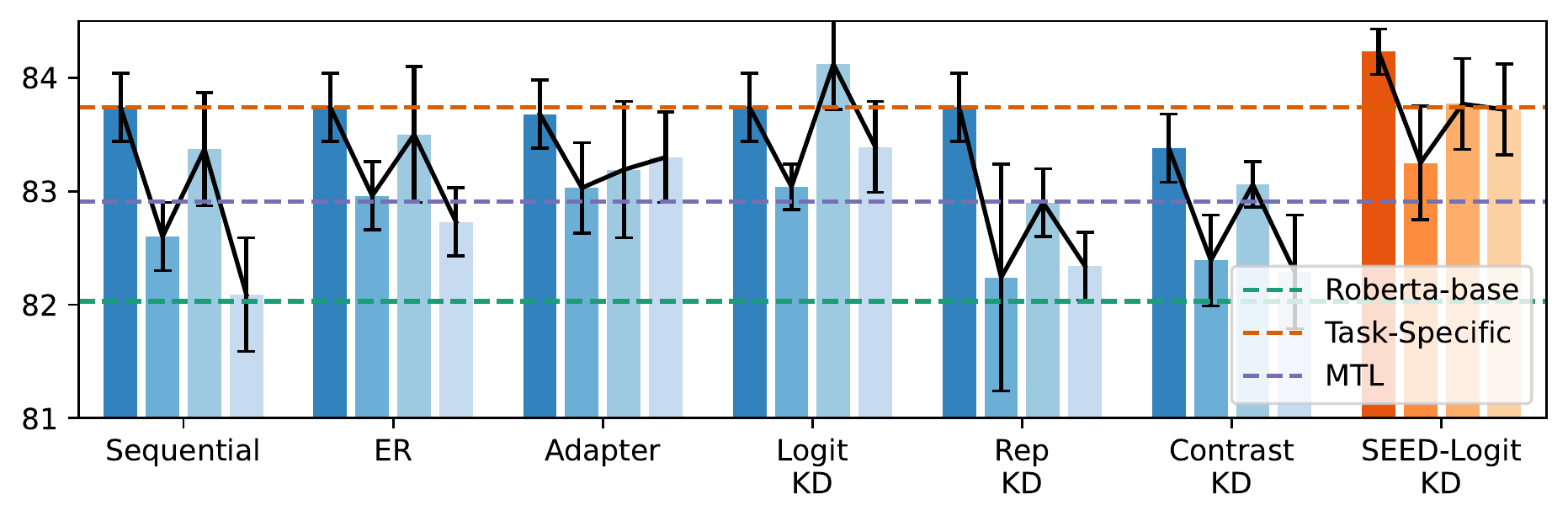}}
    \subfloat[\scriptsize{RCT-Sample}]{\includegraphics[width=0.48\linewidth]{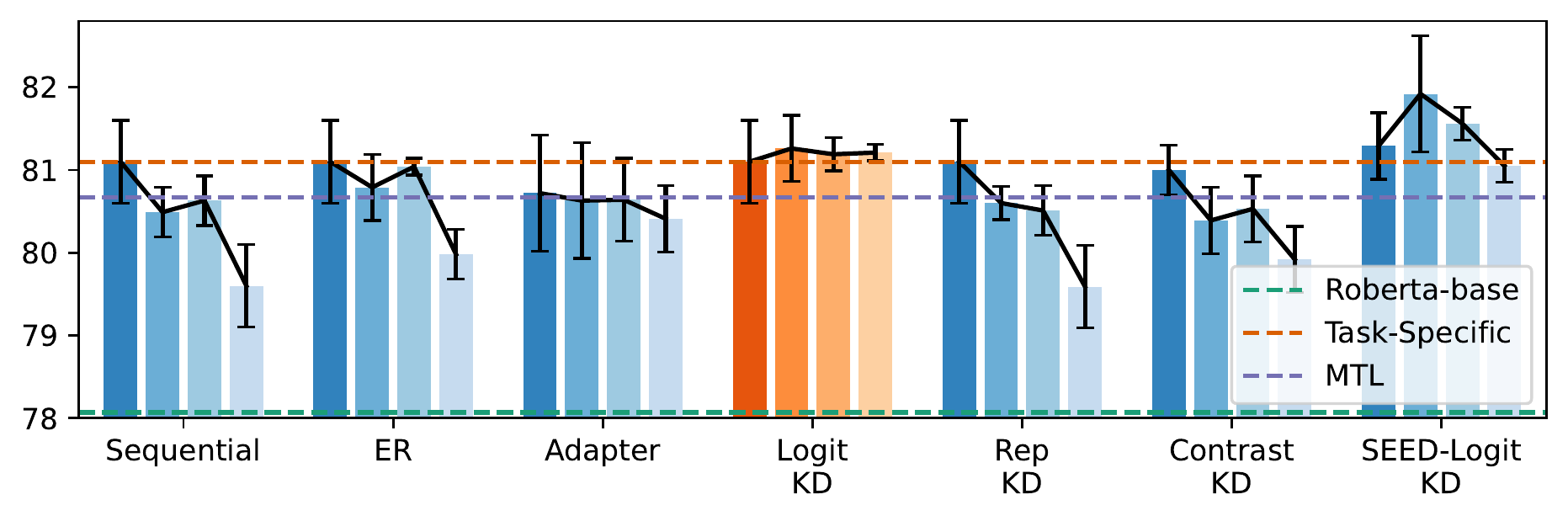}}

    \subfloat[\scriptsize{ACL-ARC}]{\includegraphics[width=0.48\linewidth]{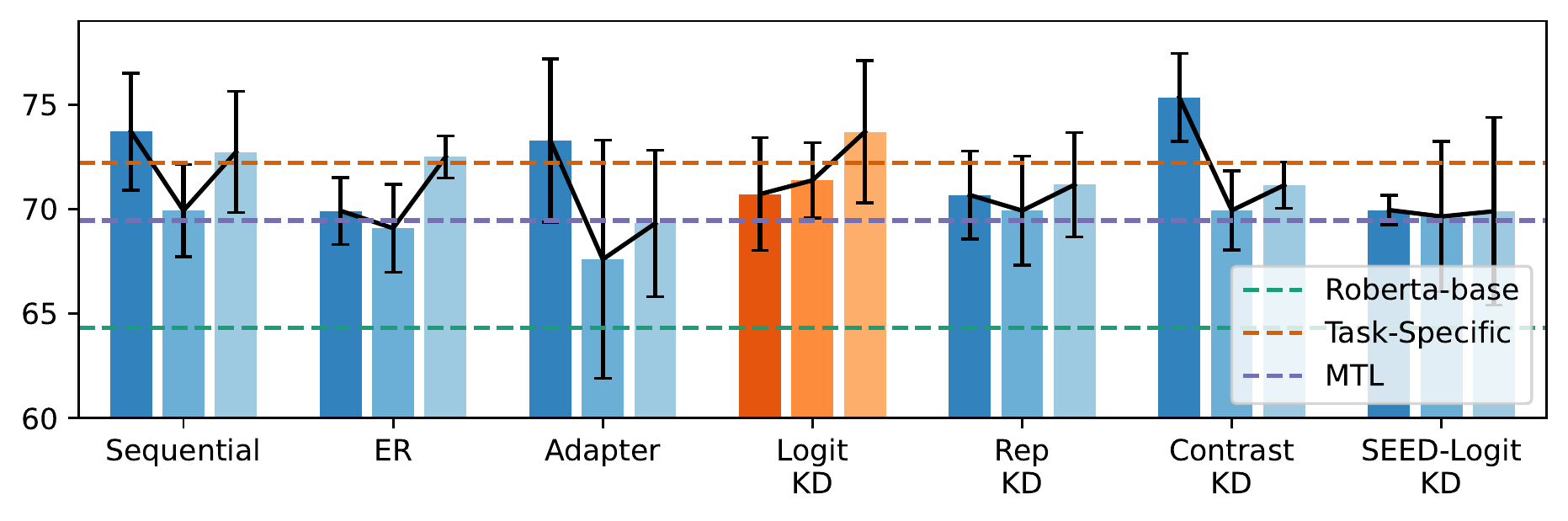}}
    \subfloat[\scriptsize{SciERC}]{\includegraphics[width=0.48\linewidth]{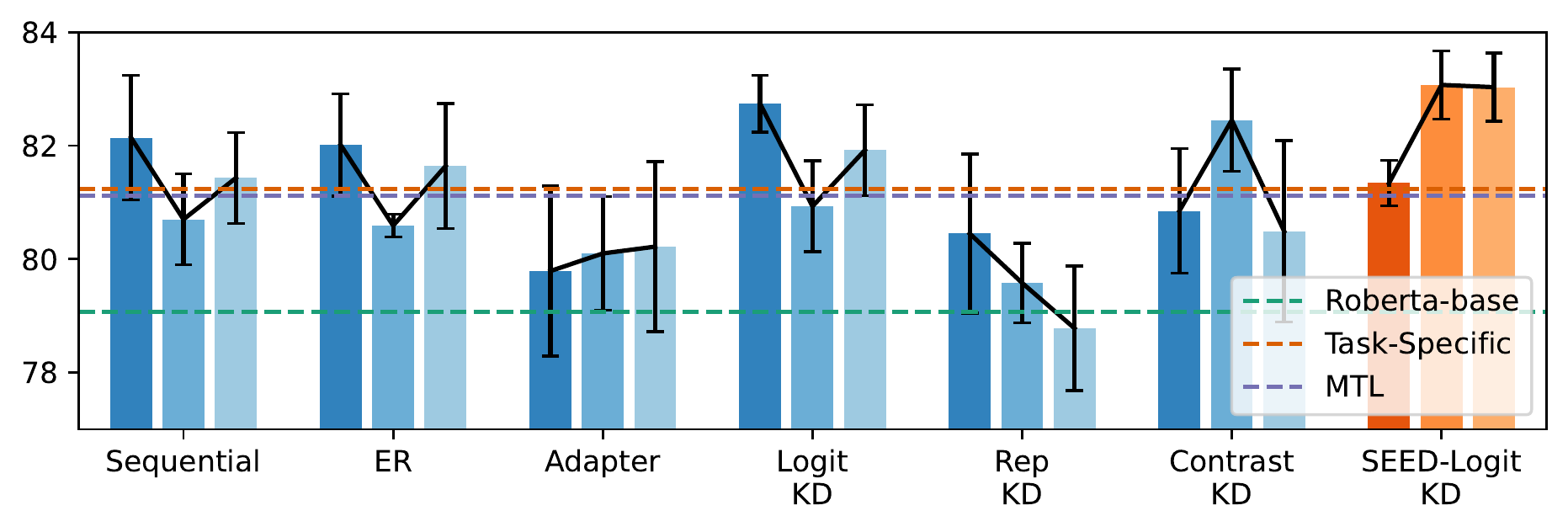}}
    \caption{\small \textbf{Performance evolution of downstream models.} Models are fine-tuned from checkpoints of lifelong pretrained LMs at different time steps $t$. For Chemprot and RCT-Sample from $D_1$, we use $t \in \{1,2,3,4\}$; while for ACL-ARC and SciERC from $D_2$, $t \in \{2,3,4\}$. Methods achieving the best performance at the end of training ($t=4$) is highlighted.}
    \label{fig:academics_curve}
    \vspace{-0.3cm}
\end{figure*}

As we introduced in Sec.~\ref{ssec:datasets}, in the domain incremental research paper stream, 
we expect a model $f_t$ to perform well on all downstream tasks $S_{1..t}$ from domains $D_{1..t}$.
In Table~\ref{tab:academics}, we report the performance of models on all downstream tasks $S_{1..T}$ fine-tuned from the final pretraining checkpoint, $f_T$. 
We visualize more complete change of downstream task performance over different time steps of pretraining (\textit{i.e.,}, $f_1, f_2,f_3,f_4$) in Fig.~\ref{fig:academics_curve}. We also report the log perplexity of masked language modeling (MLM) in Table~\ref{tab:academics} as additional information. With these results, we address the research questions below.





\paragraph{\textit{Does lifelong pretraining help retain knowledge across different domain corpora?}}
We first examine whether task-specific or lifelong pretraining improves performance over domain-specific downstream tasks. 
Comparing Task-Specific LMs with RoBERTa-base in Table~\ref{tab:academics}, we notice consistent performance improvements, especially on Biomedical and Computer Science domains ($D_1, D_2$). 
We also see Sequential Pretraining could consistently outperform RoBERTa-base. However, the comparison between Sequential Pretraining and Task Specific LMs are mixed: on $D_1, D_2, D_3$, Sequential Pretraining could outperform Task-Specific LMs only except MNER; while on the earliest biomedical domain ($D_1$), Sequential Pretraining achieves substantially lower performance. From Figure~\ref{fig:academics_curve}, we see the performance of Sequential Pretraining on Chemprot and RCT (from $D_1$) drops significantly from $t=1$ to $4$. 
The results imply lifelong pretraining allows later domains to benefit from knowledge transfer from earlier domains, but the performance on earlier domains is limited because of forgetting.


\begin{table}[]
\centering
\scalebox{0.66}{
\begin{tabular}{@{}lccccc@{}}
\toprule
$|M|$, k       & Chemprot & RCT   & ACL-ARC & SciERC & MLM-$D_{1,2}$ \\ \midrule
$100k, 10$  & 82.73    & 79.98 & 72.50   & 81.64  & 1.737/1.857     \\
$100k, 100$ & 82.06    & 78.64 & 71.97   & 81.62  & 1.599/1.789     \\
$10M, 10$  & 82.87    & 79.98 & 71.80   & 81.63  & 1.438/1.732     \\ \bottomrule
\end{tabular}
}
\caption{\small Downstream task and MLM performance of $f_T$ under different memory sizes $|M|$ and the frequency of replay $k$ (replaying every $k$ steps of training) in ER.}
\label{tab:er_tune}
\vspace{-0.2cm}
\end{table}

\paragraph{\textit{Does continual learning algorithms help retain knowledge in sequential pretraining?}}
Next, we compare different kinds of CL algorithms and investigate the effect of CL algorithms in alleviating forgetting and improving knowledge transfer. 
Table~\ref{tab:academics} shows that Online-EWC slightly improves MLM perplexity compared to  Sequential PT, but brings no improvement to the fine-tuning performance. We hypothesize that regularization directly in the parameter space as in Online-EWC is not effective when the parameter space is very high dimensional.
Adapter improves downstream task F1 scores on the bio-medical domain ($D_1$) by 1.2\% and 0.8\%, 
but does not outperform Sequential Pretraining in other domains (similarly for Simple Layer Expansion approach), 
likely because a great portion of the model is kept frozen. 

\begin{figure}[]
    \centering
    
\includegraphics[width=0.98\linewidth]{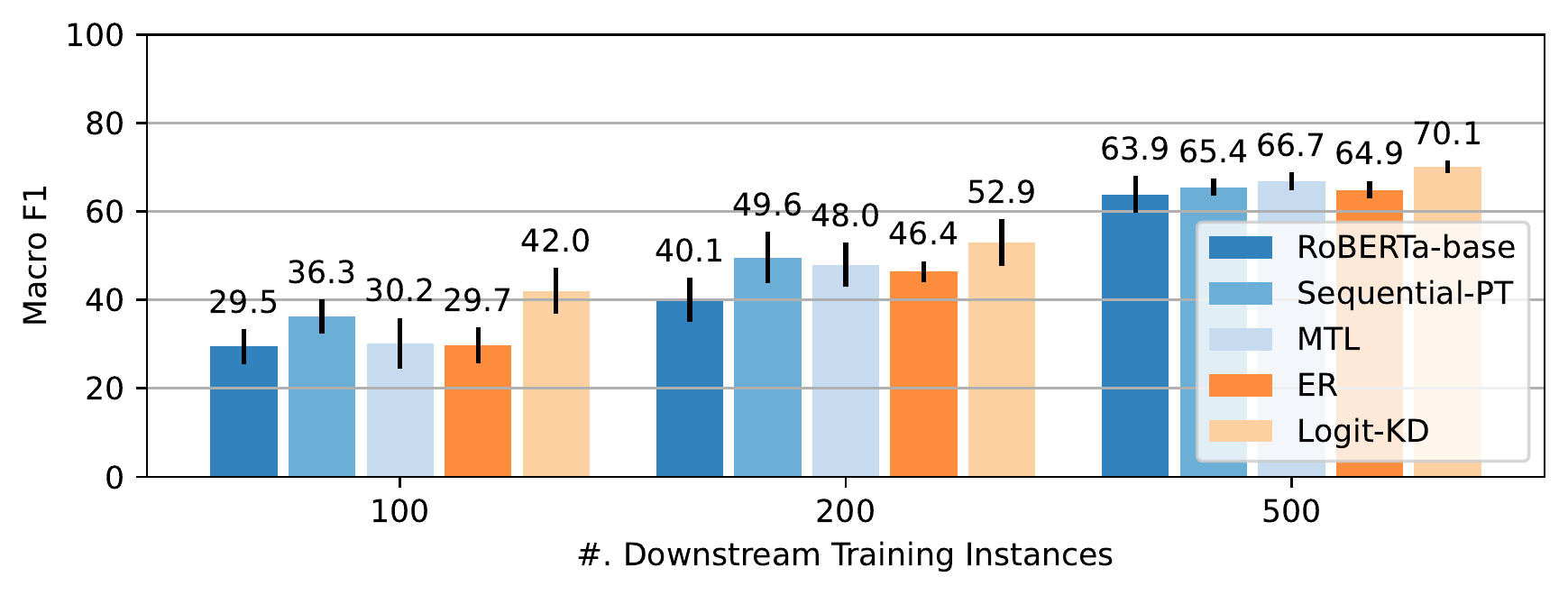}
    \caption{\small \textbf{Performance of downstream models with various number of training examples}, exemplified with SciERC. The models are fine-tuned from the final pretrained model ($f_4$).}
    \label{fig:academics_kshot_subset}
\end{figure}

In contrast, the memory-replay based approach (ER) allows training the full parameters of the model and has been shown to be highly effective in continual learning of classification tasks~\cite{Wang2019SentenceEA, chaudhry2019tiny}. However, we surprisingly find that ER could hardly improve over Sequential Pretraining except $D_1$. 
A similar pattern can be found in the MLM perplexity. 
We hypothesize that the positive effect of example replay has diminished because of the overfitting to the memory examples. Table~\ref{tab:er_tune} summarizes the effect of tuning hyperpameters in ER. When we reduce the frequency of replay (from every 10 steps to 100 steps), the MLM performance improves, which implies reduced overfitting; however, the performance of downstream task performance does not improve. When we increase the size of the memory $|M|$ from $100k$ to $10M$, the MLM perplexity also improves; still, there are still no improvements in downstream tasks. It may imply ER itself is not an effective approach for continual pretraining. 


Unlike ER, distillation approaches utilize richer information such as output logits or representation similarity to preserve past knowledge. We find either Logit KD or SEED-Logit KD to be most effective depending on the task, while Rep-KD and Contrastive-KD are less effective. The best performing distillation approach improves F1 over Sequential Pretraining on downstream tasks from $D_1$, $D_2$ at least by 1.0\%. However, performance on $D_3, D_4$, which come later in the data stream, does not improve over Sequential Pretraining, possibly because the distillation loss term 
makes the model rigid in obtaining new knowledge.

\begin{table*}[]
\centering
\scalebox{0.70}{
\begin{tabular}{@{}lccccc@{}}
\toprule
Method                                     & \#. of Forward & \#. of Backward & \#. Total   & \#. Total ($k$=10) & \ Wall Time$_{4k}$ \\ \midrule
\textit{Main results}                      &                          &                            &            &      &     \\
Sequential PT                              & $b$                      & $b$                        & $2b$       & $2b$   &  $4.0\times 10^4$ sec.  \\
ER                                         & $(1+1/k)b$               & $(1+1/k)b$                 & $(2+2/k)b$ & $2.2b$ & $4.2\times 10^4$ sec.  \\
Logit-Distill                              & $(2+2/k)b$               & $(1+1/k)b$                 & $(3+3/k)b$ & $3.3b$ &  $6.9\times 10^4$ sec.  \\
SEED-Logit-Distill                         & $(3+3/k)b$               & $(2+2/k)b$                 & $(5+5/k)b$ & $5.5b$  & $9.7\times 10^4$ sec.\\
\midrule
\textit{Additional Controlled Experiments} &                          &                            &            &        &  \\
Sequential PT$_{b^\prime=1.2b}$                     & $1.2b$                   & $1.2b$                     & $2.4b$     & $2.4b$  &  $4.4 \times 10^4$ sec. \\
ER$_{k=5}$   & $1.2b$                   & $1.2b$                     & $2.4b$     & $2.4b$  &  $4.4 \times 10^4$ sec.  \\
Sparse Logit-KD                            & $1.3b$                  & $1.1b$                    & $2.4b$     & $2.4b$  & $4.4 \times 10^4$ sec.  \\
Sparse SEED-Logit-KD$_{\backslash contrast}$         & $1.3b$                  & $1.1b$                    & $2.4b$     & $2.4b$  &  $4.8 \times 10^4$ sec. \\ \bottomrule
\end{tabular}
}
\caption{Number of forward and backward passes over PTLMs and wall clock time of different approaches. The number of forward and backwards passes are computed over visits of $b$ batches from the training data stream, where $k$ is the frequency of replay. The wall clock time is calculated over 4$k$ steps of training (which is the number of training steps of a single domain in the Research Paper stream) excluding the first domain, as no replay or distillation happens while learning the first domain. In the additional controlled experiments (described in~Appendix.~\ref{apdx:controlled_computation_costs}), we control the total number of forward and backward passes of different approaches. 
}
\label{tab:forward_backward_count}
\end{table*}

\paragraph{\textit{What is the gap between lifelong pretraining and multi-task learning across all the domains?}
}
Multi-Task Learning refers to the offline training paradigm, which retrain PTLMs over all corpora ($D_{1..t}$) each time a new corpus $D_t$ becomes available. 
We examine whether lifelong pretraining is comparable to multi-task pretraining in terms of performance. From Table~\ref{tab:academics} and Figure~\ref{fig:academics_curve}, we see Sequential Pretraining in general underperforms MTL except for the final domain. However, certain CL approaches, such as Logit-Distillation, could improve over MTL on all downstream tasks from the first and the second domain. We speculate the reason is that continual learning naturally provides a curriculum~\cite{Xu2020CurriculumLF, Shi2015RecurrentNN} to models where each individual task is easier to learn. 
The results have a positive implication that lifelong pretraining is not only more computationally efficient and requires less storage of past data, but may also improve the performance of pretraining.

\paragraph{\textit{Does lifelong pretraining make models more data efficient?}}
In Table~\ref{fig:academics_kshot_subset}, we further examine the performance of final pretrained models under different amounts of training examples. We include full results in Appendix~\ref{apdx:low_resource}. We find in general, performance improvements are more significant in the low-resource setup. 

\paragraph{\textit{Computational Costs.}} We quantify computational costs of different CL algorithms with the number of forward and backward passes in Table~\ref{tab:forward_backward_count} and present additional experiments with controlled computational costs in Appendix~\ref{apdx:controlled_computation_costs}. We find additional computational cost is necessary for performance improvement of distillation-based CL. However, it is not possible to trade performance simply by investing more computation budget with arbitrary CL algorithms. We leave detailed discussions in Appendix~\ref{apdx:controlled_computation_costs}.


\subsection{Temporal Data Stream}
\label{ssec:temporal_data_stream}

We conduct analysis on pretraining PTLM on chronologically-ordered tweet corpora, to understand whether lifelong pretraining helps adaptation to the latest data and improves temporal generalization ability. 
The results are summarized in Table~\ref{tab:twitter}.

\paragraph{\textit{Will LMs be outdated?}} 
We compare the performance of Task-Specific (2014) to the Task-Specific models pretrained on the year of downstream datasets (noted as Task-Specific (Latest)) and notice consistent improvements in downstream tasks in 2018 and 2020  (first two columns in Table~\ref{tab:twitter}). Sequential Pretraining could also outperform the Task-Specific (2014) model. It verifies that language models may get outdated over time, but the issue can be addressed by task-specific or lifelong pretraining over the latest corpora.


\begin{table}[]
\centering
\scalebox{0.607}{
\begin{tabular}{@{}lcccc@{}}
\toprule
Years                  & 2018 $(D_3)$              & 2020 $(D_4)$              & \begin{tabular}[c]{@{}c@{}}2014 $(D_1)$ \\ $\to$ 2020 $(D_4)$\end{tabular} & \begin{tabular}[c]{@{}c@{}}2016 $(D_2)$\\ $\to$ 2020 $(D_4)$\end{tabular} \\ \midrule
\multicolumn{5}{c}{Hashtag Prediction}                                                                   \\ \midrule
RoBERTa-base           & 48.08$_{\pm 1.0}$ & 56.42$_{\pm 0.2}$  & 39.31$_{\pm 2.7}$ & 42.23$_{\pm 2.7}$ \\
Sequential PT          & 56.79$_{\pm 0.5}$ & 59.85$_{\pm 0.4}$  & 44.00$_{\pm 1.1}$ & 49.87$_{\pm 1.8}$ \\
ER                     & 56.93$_{\pm 0.1}$ & 59.56$_{\pm 1.7}$  & 43.31$_{\pm 0.2}$ & 50.72$_{\pm 0.6}$ \\
Logit-KD               & \textbf{58.21$_{\pm 0.5}$} & 60.52$_{\pm 0.2}$  & 44.26$_{\pm 0.9}$ & 50.92$_{\pm 0.8}$ \\
Contrast-KD            & 57.94$_{\pm 0.4}$ & 59.54$_{\pm 0.3}$  & 45.22$_{\pm 0.1}$ & \textbf{52.14$_{\pm 1.1}$} \\
SEED-KD                & 56.87$_{\pm 0.2}$ & 59.71$_{\pm 0.2}$  & 43.39$_{\pm 0.4}$ & 49.62$_{\pm 1.0}$ \\
SEED-Logit-KD          & 57.75$_{\pm 0.4}$ & \textbf{60.74$_{\pm 0.6}$}  & \textbf{45.35$_{\pm 0.6}$} & 51.56$_{\pm 0.7}$ \\  \midrule
Task-Specific (2014)   & 56.16$_{\pm 0.6}$ & 59.59$_{\pm 0.3}$  & 44.34$_{\pm 0.6}$ & 49.26$_{\pm 0.7}$ \\
Task-Specific (Latest) & 56.61$_{\pm 0.4}$ & 59.87$_{\pm 0.6}$  & 43.44$_{\pm 0.5}$ & 49.41$_{\pm 1.1}$ \\
MTL                    & 57.89$_{\pm 0.4}$ & 59.95$_{\pm 0.3}$  & 44.04$_{\pm 0.3}$ & 50.37$_{\pm 0.3}$ \\
\midrule
\multicolumn{5}{c}{Emoji Prediction}                                                                     \\ \midrule
RoBERTa-base           & 25.71$_{\pm 0.1}$ & 24.42$_{\pm 0.2}$  & 12.02$_{\pm 0.4}$ & 13.24$_{\pm 0.2}$ \\
Sequential PT          & 29.30$_{\pm 0.1}$ & 27.69$_{\pm 0.1}$  & 14.20$_{\pm 0.2}$ & 16.08$_{\pm 1.4}$ \\
ER                     & 29.50$_{\pm 0.1}$ & 27.75$_{\pm 0.1}$  & 14.36$_{\pm 0.4}$ & 16.82$_{\pm 0.3}$ \\
Logit-KD               & 29.77$_{\pm 0.1}$ & 27.80$_{\pm 0.1}$  & 14.20$_{\pm 0.3}$ & 16.28$_{\pm 1.1}$ \\
Contrast-KD            & 29.48$_{\pm 0.2}$ & 27.72$_{\pm 0.3}$  & \textbf{14.42$_{\pm 0.3}$} & \textbf{17.52$_{\pm 0.1}$} \\
SEED-KD                & \textbf{30.12$_{\pm 0.1}$} & 27.66$_{\pm 0.1}$  & 14.36$_{\pm 0.1}$ & 16.97$_{\pm 0.4}$ \\
SEED-Logit-KD          & 29.98$_{\pm 0.1}$ & \textbf{27.84$_{\pm 0.2}$}  & 14.36$_{\pm 0.1}$ & 16.97$_{\pm 0.3}$ \\ \midrule
Task-Specific (2014)   & 28.94$_{\pm 0.0}$ & 26.98$_{\pm 0.2}$  & 13.39$_{\pm 0.2}$ & 15.14$_{\pm 0.2}$ \\
Task-Specific (Latest) & 29.06$_{\pm 0.2}$ & 27.19$_{\pm 0.1}$  & 13.00$_{\pm 0.2}$ & 14.48$_{\pm 0.3}$ \\
MTL                    & 29.52$_{\pm 0.2}$ & 27.47$_{\pm 0.0}$  & 14.07$_{\pm 0.2}$ & 16.64$_{\pm0.2}$   \\  \bottomrule
\end{tabular}
}
\caption{\small \textbf{Results on temporal data stream.} We show fine-tuning performance over years 2018 and 2020 ($D_3$, $D_4$) and the Temporal generalization from 2014 or 2016 to 2020 data ($D_1\to D_4$, $D_2 \to D_4$) on Twitter Hashtag and Emoji prediction datasets. Models are fine-tuned from the final pre-trained model $f_T$. We include full results on other years ($D_1$, $D_2$, $D_3 \to D_4$) in Appendix~\ref{apdx:full_tweet_results}.
}
\label{tab:twitter}
\end{table}

\paragraph{\textit{Does lifelong pretraining help improve the downstream model’s performance on latest data?}}
We show that downstream model's performance over later data ($D_3, D_4$) can be improved over Task-Specific models when continual learning algorithms are applied. From the first two columns of Table~\ref{tab:twitter}, 
we see Logit-KD and SEED-KD improve Hashtag prediction score over data of years 2018 and 2020. 
SEED-Logit KD further improves prediction F1 on Emoji prediction. 
Note that these findings are in contrast to the research paper stream, where CL algorithms do not improve performance in the latest domain $D_4$. The reason can be the higher similarity between domains in the tweet corpora making the knowledge transfer easier, which is further discussed in Appendix~\ref{apdx:analysis_data_streams}.

\paragraph{\textit{Does lifelong pretraining improve temporal generalization?}} 
Temporal generalization evaluates downstream performance over latest test data when fine-tuned over outdated training data. 
We show lifelong pretraining brings clear improvement to temporal generalization.
From Table~\ref{tab:twitter}, we see even Sequential Pretraining could improve over the model pretrained merely on the year 2020 data (Task-Specific (2020)) consistently. We find performance further improves with CL algorithms applied. SEED-Logit-KD performs best in general on crossyear hashtag prediction tasks. 
In crossyear emoji prediction, we find Contrast-KD and SEED-KD perform best. We also find that SEED-Logit-KD could slightly outperform Logit-KD. 

\section{Related Works}
\vspace{-0.1cm}

\paragraph{Domain and Temporal Adaptation of Language Models.} 
~\citet{Gururangan2020DontSP} study adaptation of PTLMs to domain-specific corpora. ~\citet{Arumae2020AnEI} study algorithms to mitigate forgetting in original PTLMs, but does not investigate forgetting that happens over a sequence of domains. \citet{Maronikolakis2021MultidomainPL,Rttger2021TemporalAO, luu2021time} proposes sequential pretraining over domains or emerging data, but did not investigate CL algorithms. Several recent studies have demonstrated the necessity of adapting LMs over time~\cite{Lazaridou2021PitfallsOS} while 
specifically focusing on factual knowledge~\cite{Dhingra2021TimeAwareLM, jang2021towards}.

\paragraph{Continual Learning Algorithms in NLP.}
Continual learning in NLP has mainly been studied for classification tasks. An effective approach is to utilize a number of stored past examples~\cite{dAutume2019EpisodicMI,Wang2020EfficientML}, or pseudo examples (\textit{e.g.,} the ones generated with a PTLM~\cite{Sun2020LAMOLLM,Kanwatchara2021RationalLA}). Recent extensions of the algorithm~\cite{Chuang2020LifelongLK} perform knowledge distillation with generated pseudo examples. Other lines of works focus on regularization over the sentence representations~\cite{Wang2019SentenceEA,Huang2021ContinualLF,Liu2019ContinualLF} or directly merging models in the parameter space~\cite{Matena2021MergingMW}. 
Model expansion-based approaches~\cite{Liu2019ContinualLF,Pfeiffer2021AdapterFusionNT}, including learning domain specific expert models~\cite{Gururangan2021DEMixLD}, are also actively studied.~\citet{wu2022pretrained} present a comparative study of algorithms in the context of continual fine-tuning over NLP tasks.

\section{Conclusion}
\vspace{-0.1cm}
In this paper, we formulated the lifelong language model pretraining problem and constructed two data streams associated with downstream datasets. 
We evaluated knowledge retention, adaptation to the latest data, and temporal generalization ability of continually pretrained language models. 
Our experiments show distillation-based approaches being most effective in these evaluation setups. A limitation of the work is that it has not been fully addressed whether there exists a variant of distillation-based CL approach that consistently outperforms Logit-KD. Based on the current observation, we conclude the performance of different KD approaches for CL is highly task-dependent.
It asks for more future works into continual learning algorithms within the proposed problem setup.


\bibliography{main}
\bibliographystyle{acl_natbib}

\appendix
\clearpage
\newpage

\section{Detailed Experiment Settings}
\label{sec:hyps}
We use a linearly decreasing learning rate initialized with 5e-4 on the research paper stream and 3e-4 on the tweet stream. On the research paper stream, we train the model for 8,000 steps in the first task, and 4,000 steps in the subsequent tasks. On the tweet stream, we train the model for 8,000 steps in all tasks. We hold out 128,000 sentences from each corpus to evaluate MLM performance. As the size of pretraining corpora is large, during training, each training example is visited only once. We use the masked language modeling perplexity over held-out validation sets of the pretraining corpora as the metrics for hyperparameter tuning. 
Common hyperparameters such as learning rate and batch sizes are tuned with Task-specific models with the first task. Hyperparameters that are specific to continual learning algorithms, such as the scale of the distillation loss, is tuned using the first two domains in the stream according to the MLM performance over validation sets. The weight of the distillation term $\alpha$ is set as 1.0 for logit distillation and 0.1 for other distillation algorithms. By default, we replay or perform distillation with a mini-batch of examples from the replay memory every 10 training steps in ER and Distillation-based CL approaches. We use the huggingface transformers library \url{https://github.com/huggingface/transformers} for implementation.


\section{Low-Resource Fine-Tuning}
\label{apdx:low_resource}

\begin{figure}[]
    \centering
    \subfloat[\scriptsize{Chemprot}]{\includegraphics[width=0.94\linewidth]{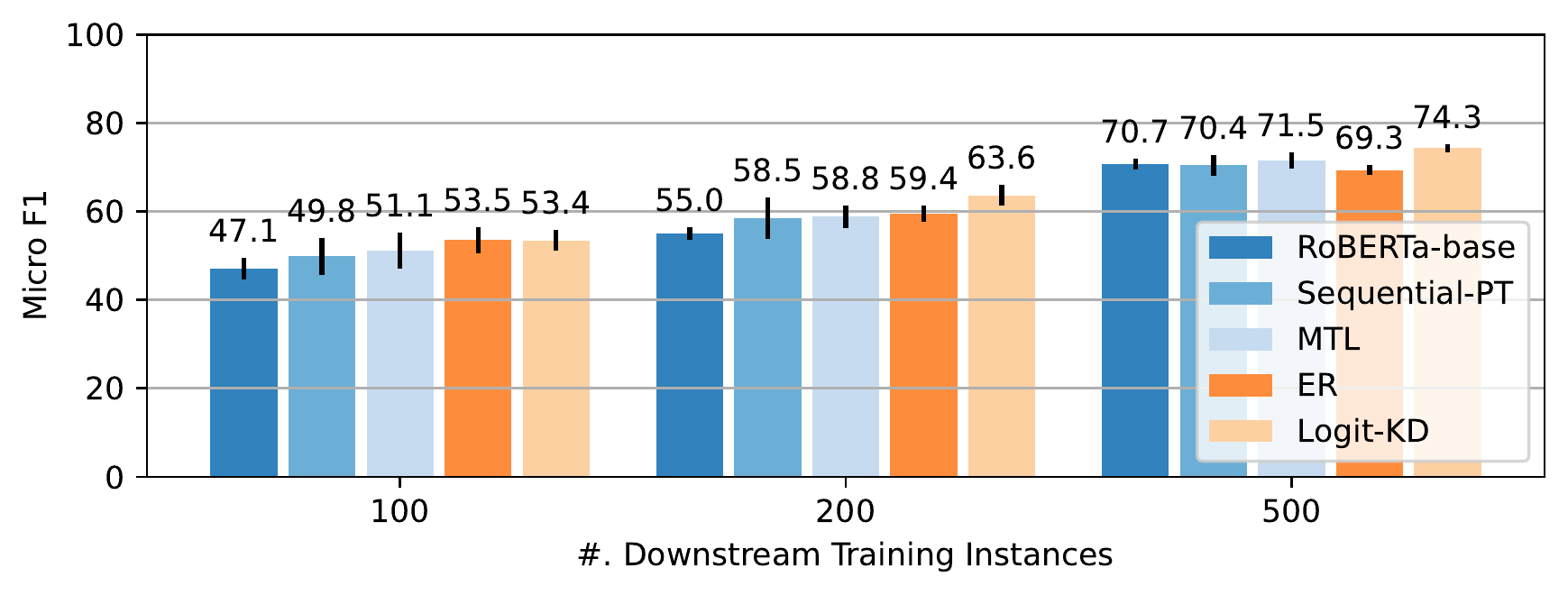}}
    
    
    
    \subfloat[\scriptsize{SciERC}]{\includegraphics[width=0.94\linewidth]{acl2022/figs/academics_kshots/kshot_sciie.pdf}}
    \caption{\small Performance of downstream models with various number of training examples. The models are fine-tuned from the final pretrained model ($f_4$).}
    \label{fig:academics_kshot}
\end{figure}

Figure~\ref{fig:academics_kshot} summarizes the performance of fine-tuned models from the final model checkpoint ($t=4$) using different amount of downstream training examples. We see on Chemprot and SciERC, the benefit of Sequential Pretraining over RoBERTa-base is more significant in low-resource fine-tuning setups. 
Whenever Seqential Pretraining outperforms RoBERTa-base, we notice Logit-KD could further improve over Sequential Pretraining.

\section{Full Results over the Tweet Stream}
\label{apdx:full_tweet_results}
\begin{table}[]
\centering
\scalebox{0.66}{
\begin{tabular}{@{}lcccc@{}}
\toprule
Task          & 2014                 & 2016                 & 2018                 & 2020                 \\ \midrule
\multicolumn{5}{c}{Hashtag Prediction}                                                      \\ \midrule
RoBERTa-base  & 56.65$_{\pm 0.6}$            & 45.50$_{\pm 2.1}$            & 48.08$_{\pm 1.0}$            & 56.42$_{\pm 0.2}$            \\
Sequential PT        & 59.00$_{\pm 0.1}$            & 54.28$_{\pm 0.3}$            & 56.79$_{\pm 0.5}$            & 59.85$_{\pm 0.4}$            \\
ER            & 59.00$_{\pm 0.1}$            & 54.90$_{\pm 0.2}$            & 56.93$_{\pm 0.1}$            & 59.56$_{\pm 1.7}$            \\
Adapter       &     58.76$_{\pm 0.7}$         &         52.55$_{\pm 1.5}$             &           54.34$_{\pm1.7}$           &          59.01$_{\pm 1.0 }$            \\
Logit-KD      &   60.93$_{\pm 0.5}$        &        55.96$_{\pm 0.2}$              &            58.21$_{\pm 0.5}$          &        60.52$_{\pm 0.2}$              \\
Rep-KD        &      60.47$_{\pm 0.1}$           &      51.77$_{\pm 2.6}$                &              55.79$_{\pm 1.4}$        &          59.80$_{\pm 0.2}$ \\
Contrast-KD   &       60.72$_{\pm 0.6}$         &      55.85$_{\pm 0.0}$              &         57.94$_{\pm 0.4}$         &     59.54$_{\pm 0.3}$                 \\
SEED-KD & 58.82$_{\pm 0.4}$ & 54.55$_{\pm 0.5}$ & 56.87$_{\pm 0.2}$ & 59.71$_{\pm 0.2}$ \\
SEED-Logit-KD &        61.28$_{\pm 0.2}$           &            55.59$_{\pm 0.5}$          &                57.75$_{\pm 0.4}$      &          60.74$_{\pm 0.6}$            \\
Task-Specific (2014) & 61.62$_{\pm 0.3}$            & 55.38$_{\pm 0.6}$            & 56.16$_{\pm 0.6}$            & 59.59$_{\pm 0.3}$            \\
Task-Specific (Latest) & 59.91$_{\pm 0.3}$            & 55.47$_{\pm 1.0}$            & 56.61$_{\pm 0.4}$            & 59.87$_{\pm 0.6}$            \\
MTL           & 60.51$_{\pm 0.3}$            & 55.16$_{\pm 1.6}$            & 57.89$_{\pm 0.4}$            & 59.95$_{\pm 0.3}$            \\ \midrule
               \multicolumn{5}{c}{Emoji Prediction}                                                      \\ \midrule
RoBERTa-base  & 28.73$_{\pm 0.2}$            & 26.86$_{\pm 0.2}$            & 25.71$_{\pm 0.1}$            & 24.42$_{\pm 0.2}$            \\
Sequential PT       & 32.69$_{\pm 0.2}$            & 30.55$_{\pm 0.3}$            & 29.30$_{\pm 0.1}$            & 27.69$_{\pm 0.1}$            \\
ER            & 32.88$_{\pm 0.2}$            & 30.52$_{\pm 0.2}$            & 29.50$_{\pm 0.1}$            & 27.75$_{\pm 0.1}$            \\
Adapter       &      32.15$_{\pm 0.2}$         &        29.85$_{\pm 0.0}$         &            28.72$_{\pm 0.0}$          &     26.80$_{\pm 0.3}$             \\
Logit-KD      & 33.08$_{\pm 0.3}$            & 30.88$_{\pm 0.1}$            & 29.77$_{\pm 0.1}$            & 27.80$_{\pm 0.1}$            \\
Rep-KD        &      32.71$_{\pm 0.2}$         &        30.51$_{\pm 0.2}$              &        29.45$_{\pm 0.1}$          &        27.27$_{\pm 0.2}$          \\
Contrast-KD   & 32.90$_{\pm 0.1}$            & 31.01$_{\pm 0.1}$            & 29.48$_{\pm 0.2}$            & 27.72$_{\pm 0.3}$            \\
SEED-KD & 32.91$_{\pm 0.1}$ & 30.84$_{\pm 0.3}$ & 30.12$_{\pm 0.1}$ & 27.66$_{\pm 0.1}$ \\
SEED-Logit-KD &      33.28$_{\pm 0.1}$          &      31.17$_{\pm 0.1}$                &           29.98$_{\pm 0.1}$       &       27.84$_{\pm 0.2}$               \\
Task-Specific (2014) & 33.37$_{\pm 0.2}$            & 30.54$_{\pm 0.3}$            & 28.94$_{\pm 0.0}$            & 26.98$_{\pm 0.2}$            \\
Task-Specific (Latest) & 32.31$_{\pm 0.0}$            & 29.83$_{\pm 0.5}$            & 29.06$_{\pm 0.2}$            & 27.19$_{\pm 0.1}$            \\
MTL           &  32.78$_{\pm 0.1}$  & 30.54$_{\pm 0.0}$ & 29.52$_{\pm 0.2}$ &  27.47$_{\pm 0.0}$ \\  \bottomrule
\end{tabular}
}
\caption{\small Full performance on Twitter Hashtag prediction and Emoji prediction, fine-tuned from the pre-trained model in the final time step.}
\label{tab:twitter_full}
\end{table}


\begin{table}[]
\centering
\scalebox{0.66}{
\begin{tabular}{@{}lccc@{}}
\toprule
Task          & 2014 $\to$ 2020 & 2016 $\to$ 2020 & 2018 $\to$ 2020 \\ 
\midrule
\multicolumn{4}{c}{Crossyear Hashtag Prediction}                        \\
\midrule
RoBERTa-base  & 39.31$_{\pm 2.7}$     & 42.23$_{\pm 2.7}$     & 37.19$_{\pm 2.1}$     \\
Sequential PT       & 44.00$_{\pm 1.1}$     & 49.87$_{\pm 1.8}$     & 46.63$_{\pm 0.9}$     \\
ER            & 43.31$_{\pm 0.2}$     & 50.72$_{\pm 0.6}$     & 46.27$_{\pm 0.4}$     \\
Adapter       &    42.61$_{\pm 0.5}$           &        48.00$_{\pm 1.6}$       &   42.63$_{\pm 0.9}$            \\
Logit-KD      & 44.26$_{\pm 0.9}$     & 50.92$_{\pm 0.8}$     & 46.84$_{\pm 1.0}$     \\
Rep-KD        &     42.48$_{\pm 0.2}$      &        50.38$_{\pm 1.5}$       &       42.23$_{\pm 0.2}$        \\
Contrast-KD   &      45.22$_{\pm 0.1}$       &       52.14$_{\pm 1.1}$         &            47.47$_{\pm 0.8}$   \\
SEED-KD & 43.39$_{\pm 0.4}$  & 49.62$_{\pm 1.0}$ & 46.37$_{\pm 0.8}$ \\
SEED-Logit-KD &   45.35$_{\pm 0.6}$      &      51.56$_{\pm 0.7}$     &         47.74$_{\pm 0.3}$      \\
Task-Specific (2014) &      44.34$_{\pm 0.6}$         &         49.26$_{\pm 0.7}$      &        45.09$_{\pm 0.7}$       \\
Task-Specific (2020) &      43.44$_{\pm 0.5}$          &           49.41$_{\pm 1.1}$                &          44.34$_{\pm 0.4}$        \\
- \textit{4x steps}& 44.34$_{\pm 0.6}$     & 51.78$_{\pm 0.7}$     & 44.69$_{\pm 0.7}$     \\
MTL           & 44.04$_{\pm 0.3}$     & 50.37$_{\pm 0.3}$     & 44.31$_{\pm 0.0}$     \\ 
\midrule
\multicolumn{4}{c}{Crossyear Emoji Prediction}                          \\ \midrule
RoBERTa-base  & 12.02$_{\pm 0.4}$     & 13.24$_{\pm 0.2}$     & 18.67$_{\pm 0.1}$     \\
Sequential PT       & 14.20$_{\pm 0.2}$     & 16.08$_{\pm 1.4}$     & 21.06$_{\pm 0.9}$     \\
ER            & 14.36$_{\pm 0.4}$     & 16.82$_{\pm 0.3}$     & 21.57$_{\pm 0.1}$     \\
Adapter       &     13.53$_{\pm 0.2}$       &   15.68$_{\pm 0.3}$        &      20.64$_{\pm 0.1}$         \\
Logit-KD      & 14.20$_{\pm 0.3}$     & 16.28$_{\pm 1.1}$     & 21.29$_{\pm 1.0}$     \\
Rep-KD        &    13.89$_{\pm 0.1}$       &        16.03$_{\pm 0.3}$       &           20.86$_{\pm 0.2}$    \\
Contrast-KD   &     14.42$_{\pm 0.3}$       &       17.52$_{\pm 0.1}$        &      21.43$_{\pm 0.1}$         \\
SEED-KD & 14.36$_{\pm 0.1}$ & 16.97$_{\pm 0.4}$ & 21.88$_{\pm 0.3}$ \\
SEED-Logit-KD & 14.36$_{\pm 0.1}$    &     16.97$_{\pm 0.3}$      &         21.62$_{\pm 0.1}$      \\
Task-Specific (2014) &    13.39$_{\pm 0.2}$        &        15.14$_{\pm 0.2}$                   &       20.79$_{\pm 0.3}$              \\
Task-Specific (2020) & 13.00$_{\pm 0.2}$     & 14.48$_{\pm 0.3}$     & 19.30$_{\pm 0.2}$     \\
- \textit{4x steps}& 12.90$_{\pm 0.4}$     & 14.85$_{\pm 0.3}$     & 19.83$_{\pm 0.2}$     \\
MTL           &     14.07$_{\pm 0.2}$       &       16.64$_{\pm0.2}$    &       20.94$_{\pm 0.7}$   \\ 
\bottomrule
\end{tabular}
}
\caption{\small Temporal generalization performance on Twitter Hashtag prediction datasets fine-tuned from the final pre-trained model. \texttt{Year 1}$\to$\texttt{Year 2} indicates the hashtag prediction model is fine-tuned on data in year \texttt{Year 1}, and evaluated on test data in \texttt{Year 2}.}
\vspace{-0.3cm}
\label{tab:twitter_tg_full}
\end{table}

Tables~\ref{tab:twitter_full} and~\ref{tab:twitter_tg_full} summarize full results over the Tweet stream. Compared to the table~\ref{tab:twitter} in the main text, we add downstream performance over data from years 2014 and 2016 ($D_1$, $D_2$), and temporal generalization from year 2014 to 2020 ($D_1 \to D_4$). 

\section{Dataset Details}
\label{apdx:dataset_details}
The research paper stream consists of full text of 6.6M, 12.1M, 7.8M, and 7.5M research papers from the S2ORC~\cite{Lo2020S2ORCTS} dataset. We evaluate downstream fine-tuning performance on two in-domain datasets for each research area:
Chemprot relation exaction dataset~\cite{Vindahl2016ChemProt30AG} and RCT abstract sentence role labeling dataset ~\cite{Dernoncourt2017PubMed2R} for the bio-medical domain; 
ACL-ARC citation intent classification dataset~\cite{Jurgens2018MeasuringTE} and SciERC relation extraction dataset~\cite{Luan2018MultiTaskIO} for the computer science domain; 
relation extraction over Synthesis procedures~\cite{mysore2019materials} and named entity recognition over material science papers (MNER)~\cite{olivetti2020data} for material science domain; keyphrase classification and hyponym classification after filtering out physics papers for the physics domain~\cite{augenstein2017semeval}. We report micro-averaged F1 on Chemprot, RCT, MNER datasets following the evaluation metrics in the original work, and report macro-averaged F1 on all other datasets. We use the official data splits for all datasets except for RCT, where we employ a low-resource training setup following~\citet{Gururangan2020DontSP}. 

The pretraining corpora for the tweet stream consist of 25M tweets in each year. 
For downstream tasks, we use a separate set of 1M tweets from each year to construct multi-label hashtag prediction~\cite{Gong2016HashtagRU} datasets and single-label emoji prediction datasets~\cite{Barbieri2018SemEval2T}. We replace user names to special tokens. For Hashtag prediction, the label space consists of tweets containing 200 most frequent hashtags in each year. We independently sample 500 tweets per label (hashtag) as training, validation and test sets, which results 10$k$ examples in each of the data splits. For emoji prediction, we construct 20-way single-label emoji prediction datasets for each year following~\citet{Barbieri2018SemEval2T} with the 1M held out tweets. We sample 5,000 tweets per emoji in each split, resulting in balanced datasets of the same size as the hashtag prediction datasets.

\section{Details of Continual Learning Algorithms}
\label{apdx:cl}
\subsection{Contrastive Distillation}
During continual pretraining, in addition to the language model pretraining objective, we add a unsupervised contrastive learning objective, namely the SimCSE~\cite{Gao2021SimCSESC} objective, so that the similarity in the sentence representation better reflects the semantic similarity in the sentence. 
We use the $l^2$-normalized representation of the start-of-sequence token at the final layer as the sentence representation, noted as $\bm{h}$. Then, we distill the intra-batch representational similarity from the previous model $f_{t-1}$ to the current model $f_t$. Given a mini-batch of $N$ examples $\bm{x}$, we compute the representational dot-product similarity matrix between normalized sentence representations $\bm{h}$ between each pair of examples with $f_{t-1}$ and $f_t$, noted as $\mathbf{B}^{t-1}$ and $\mathbf{B}^{t}$, where each element $\mathbf{B}_{ij}$ is, 
\begin{equation}
    \label{eq:normalize}
    \mathbf{B}_{ij} = \frac{\exp(\bm{h}_i \cdot \bm{h}_j / \tau)}{\sum_{k=1..N}\exp(\bm{h}_i \cdot \bm{h}_k / \tau)}
\end{equation}
where $\tau$ is a temperature hyperparameter. We specify a temperature $\tau_t=0.05$ for the teacher model $f_{t-1}$ and a temperature $\tau_s$ for the student model $f_t=0.01$. We compute the cross-entropy between $\mathbf{B}^{t-1}$ and $\mathbf{B}^{t}$ as the distillation loss,
\begin{equation}
    \label{eq:contrastive_distill}
    \ell_{\text{distill}} = - \frac{1}{N} \sum_{i=1}^{N} \sum_{j=1}^{N} \mathbf{B}_{ij}^{t-1}\log \mathbf{B}_{ij}^{t}
\end{equation}

\subsection{SEED Distillation}
SEED distillation proposed by~\cite{Fang2021SEEDSD} has a similar spirit as the contrastive distillation with differences in the examples used for computing similarity matrices computes. The algorithm distills representational similarity \textit{between the batch and a large set of other examples}, maintained in an example queue $Q$.
As the number of target examples $K$ can be much larger than the batch size, it allows distillation of richer information by regularizing similarities. During pretraining, the method maintains a fixed-size queue $Q$ to cache examples from the current domain $D_t$. Given a mini-batch of training examples $\bm{x}$, it computes cosine similarity between each pair of examples within the batch $\bm{x}$ and $Q$ with $f_{t-1}$ and $f_{t}$, resulting in two similarity matrices $\mathbf{B}^{t-1}$, $\mathbf{P}^y \in \mathbb{R}^{|B|\times |Q|}$. Similar to the contrastive distillation, the distillation loss is the cross-entropy between two similarity matrices $\mathbf{B}^{t-1}$ and $\mathbf{B}^{t}$ computed in the same way as Eq.~\ref{eq:contrastive_distill}.

\section{Analysis and Controlled Experiments of Computational Costs}
\label{apdx:controlled_computation_costs}

Computational cost is a crucial matter for online continual learning systems. In this section, we analyze the computational costs of continual learning algorithms and perform controlled experiments of computational costs.

We quantify computational costs with the total number of \textit{forward ($C_f$) and backward ($C_b$)} computations ($C=C_f+C_b$) over the PTLMs, which is easy to control; in practice, we find the wall clock time of training was approximately linear to $C$. We summarize the number of forward and backward passes and the wall clock time of training in Table~\ref{tab:forward_backward_count}. In the visit of $b$ batches from the training stream, Sequential PT performs $b$ forward and backward passes respectively over the PTLM, resulting in $C=2b$. Experience replay further replays 1 batch of examples every $k$ steps over the training stream, which results in $C=(2+2/k)b$. In our main experiments, $r$ is set to 10 (Sec.~\ref{ssec:mem_replay_methods}). Logit-Distill and Rep-Distill require one additional forward pass over a frozen PTLM to compute the target of distillation, resulting in $C=(3+3/k)b$. Distillation algorithms that perform contrastive learning with SimCSE (\textit{i.e.} SEED-Distill and SEED-Logit-Distill) additionally require one forward and backward pass using the same batch of examples with different dropout masks. Therefore, for SEED-Logit-Distill, $C=(5+5/k)b$.

To control the number of forward and backward passes, we present approaches to compensate the lower computation costs compared to Distillation algorithms and one approach to shrink the computational cost of distillation algorithms: (1) for Sequential PT, we train the models for 1.2 times more steps so that $C=2.4b$, noted as Sequential PT$_{b^\prime=1.2b}$; (2) for ER, we increase the replay frequency $k$ to 5 from the default setup 10, so that $C=2.4b$. We also decrease the cost of Logit-KD and SEED-Logit-KD by reducing the frequency of distillation from every 1 batch to every $r^\prime=$10 steps, while still replaying and distilling knowledge over 1 batch of memory examples every 10 training steps. This results in $C_f=(1+2/k+1/k^\prime)b$ and $C_b=(1+1/k)b$, where $C=2.4b$ when both $r$ and $r^\prime$ are 10. The approach is referred to as Sparse Logit-KD. Finally, for SEED-Logit-KD, we remove the SimCSE loss from training and perform sparse distillation similar to Sparse-Logit-KD, which also results in $C=2.4b$.

The performance of the models is presented in Table~\ref{tab:controlled_results}. We notice that at the end of pretraining, increasing the number of training steps in Sequential PT by 1.2 times does not lead to performance boost on the latest domain ($D_4$), while the performance over tasks from earlier domains (Chemprot, ACL-ARC, SciERC) slightly dropped, possibly due to increased forgetting. For ER, we notice replaying only slightly more frequently (ER$_{k=5}$) than the default setup ($k$=10) greatly increased the perplexity of MLM, implying significantly increased overfitting to the memory; while the performance differences of downstream tasks compared to the default ER is mixed. When we decrease the replay frequency of distillation, the performance on Logit-KD and SEED-KD also decreased and does not outperform ER.

The results show additional computation costs can be necessary for continual learning algorithms such as Logit-KD and SEED-Logit-KD. However, the results also show that there is no simple trade-off between computational cost and performance. We have seen that it is not always beneficial to increase the number of training steps over the emerging data, as it increases forgetting in earlier domains. Similarly, increasing the frequency of replay may lead to significant overfitting to the replay memory. Investigating into more effective continual learning algorithms, despite increased computation costs, allows us to obtain performance improvement that cannot be simply traded with more computation with arbitrary continual learning algorithms.  We leave more thorough studies into this topic as future work.

\begin{table}[]
\centering
\scalebox{0.66}{
\begin{tabular}{@{}lcccc@{}}
\toprule
Domain        & \multicolumn{2}{c}{Biomedical} & \multicolumn{2}{c}{Computer Science} \\
\cmidrule(r){1-1} \cmidrule(lr){2-3} \cmidrule(lr){4-5}  
Dataset       & Chemprot      & RCT-Sample     & ACL-ARC           & SciERC           \\ \midrule
RoBERTa-large & 84.39$_{\pm0.7}$     & 80.76$_{\pm0.7}$      & 72.20$_{\pm3.2}$         & 83.02$_{\pm0.8}$        \\
Naive         & 85.43$_{\pm0.6}$     & 81.10$_{\pm0.5}$      & 73.44$_{\pm2.0}$         & 82.88$_{\pm0.7}$        \\
ER            & 85.42$_{\pm0.2}$     & 81.30$_{\pm0.4}$      & 71.51$_{\pm2.5}$         & 83.22$_{\pm0.5}$        \\
Logit-KD      & \textbf{86.18$_{\pm0.7}$}     & \textbf{81.93$_{\pm0.7}$}      & 72.10$_{\pm2.0}$         & 83.23$_{\pm0.6}$        \\
SEED-Logit-KD  &      85.98$_{\pm 0.4}$               &         81.34$_{\pm 0.4}$             &         74.02$_{\pm 3.0}$            &   \textbf{83.89$_{\pm 0.6}$}
\\
Task-Specific        & 85.99$_{\pm0.3}$     & 82.02$_{\pm0.6}$      & \textbf{76.07$_{\pm1.0}$}         & 82.91$_{\pm0.9}$        \\ \bottomrule
\end{tabular}}
\caption{Results on the Research Paper stream with RoBERTa-large as the base model.}
\label{tab:roberta_large}
\end{table}

\begin{figure*}[!t]
    \centering
    \subfloat[\scriptsize{Chemprot}]{\includegraphics[width=0.48\linewidth]{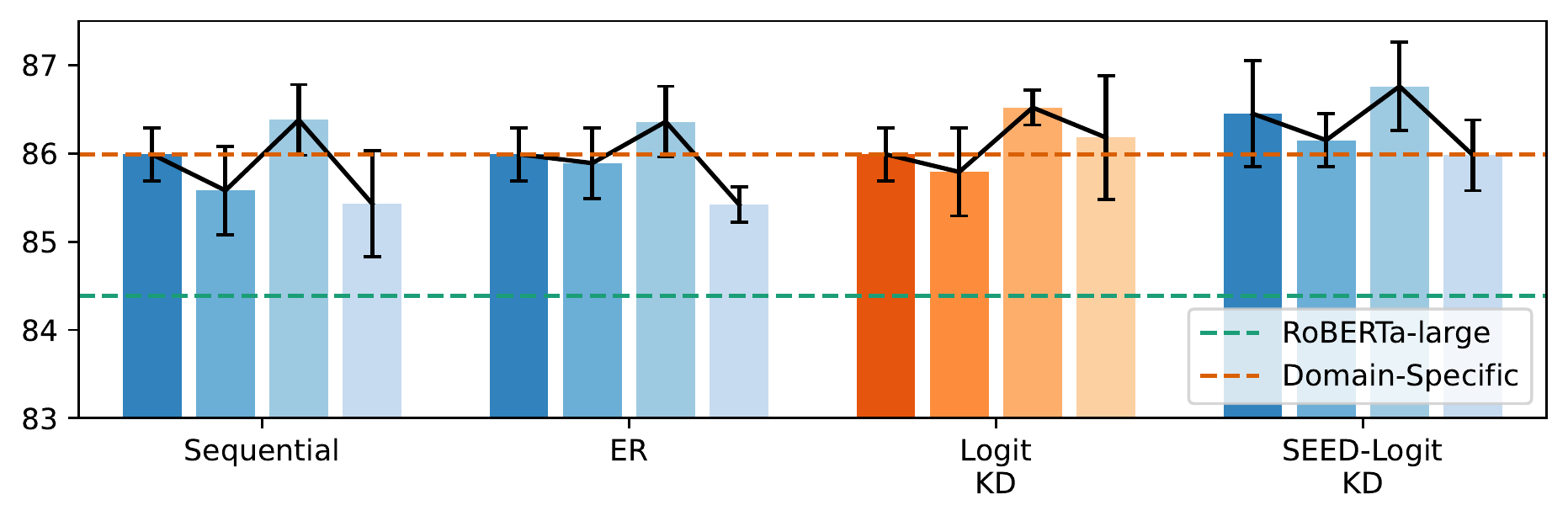}}
    \subfloat[\scriptsize{RCT-Sample}]{\includegraphics[width=0.48\linewidth]{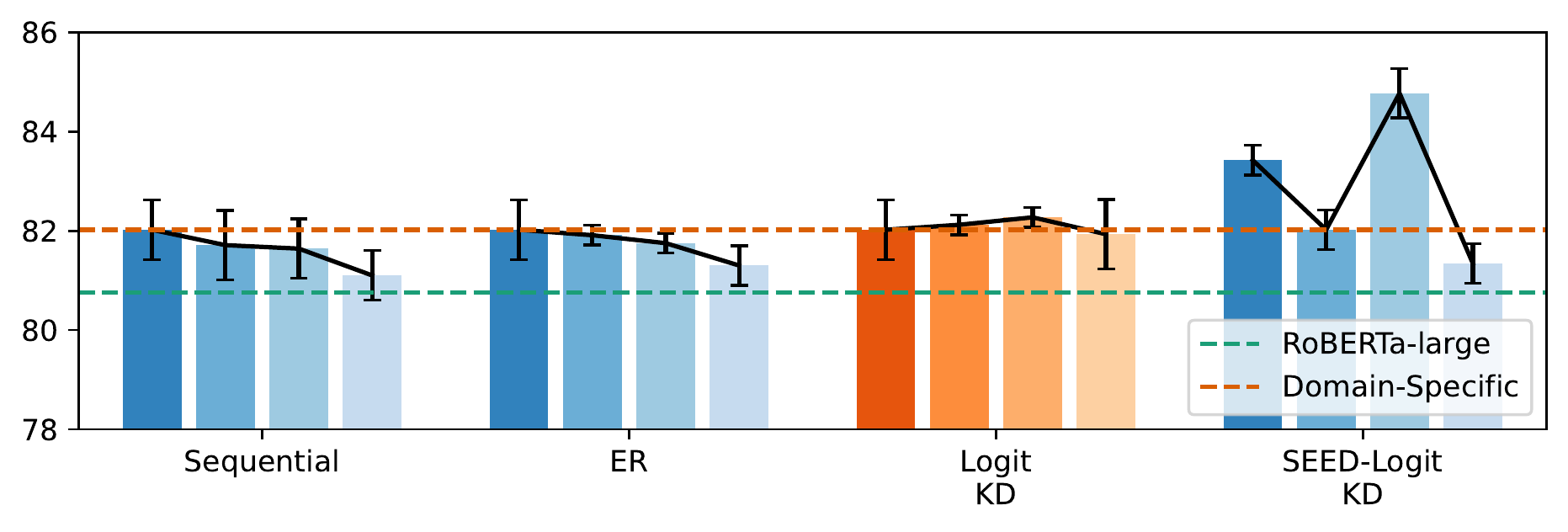}}

    \subfloat[\scriptsize{ACL-ARC}]{\includegraphics[width=0.48\linewidth]{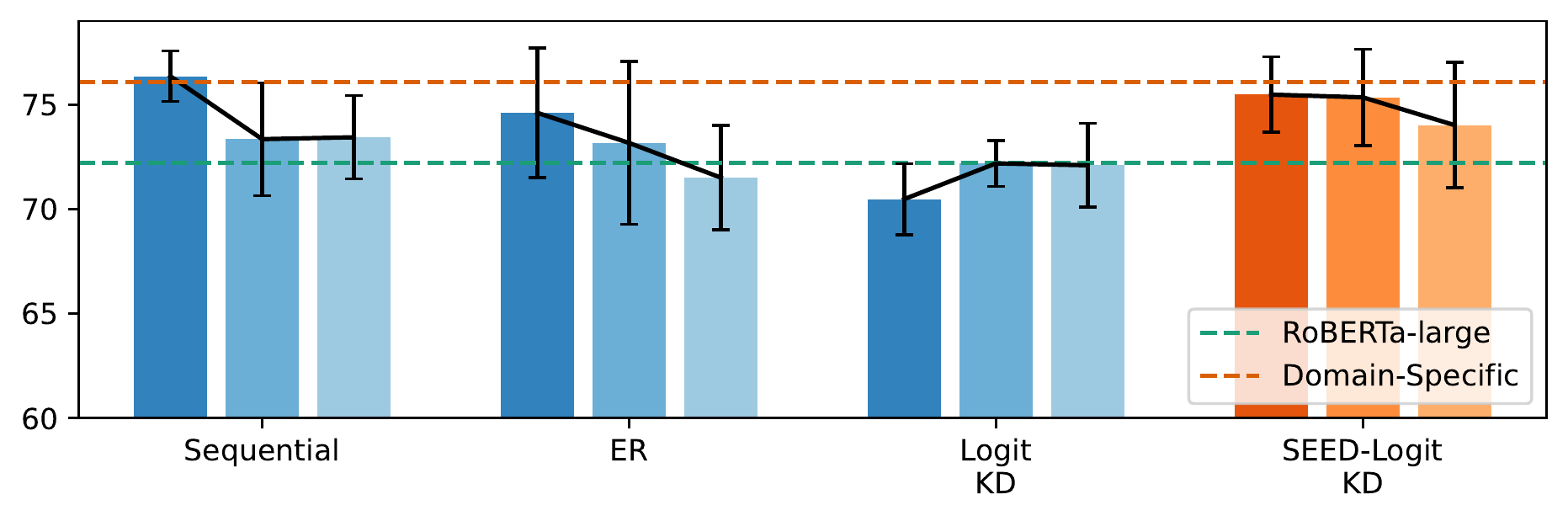}}
    \subfloat[\scriptsize{SciERC}]{\includegraphics[width=0.48\linewidth]{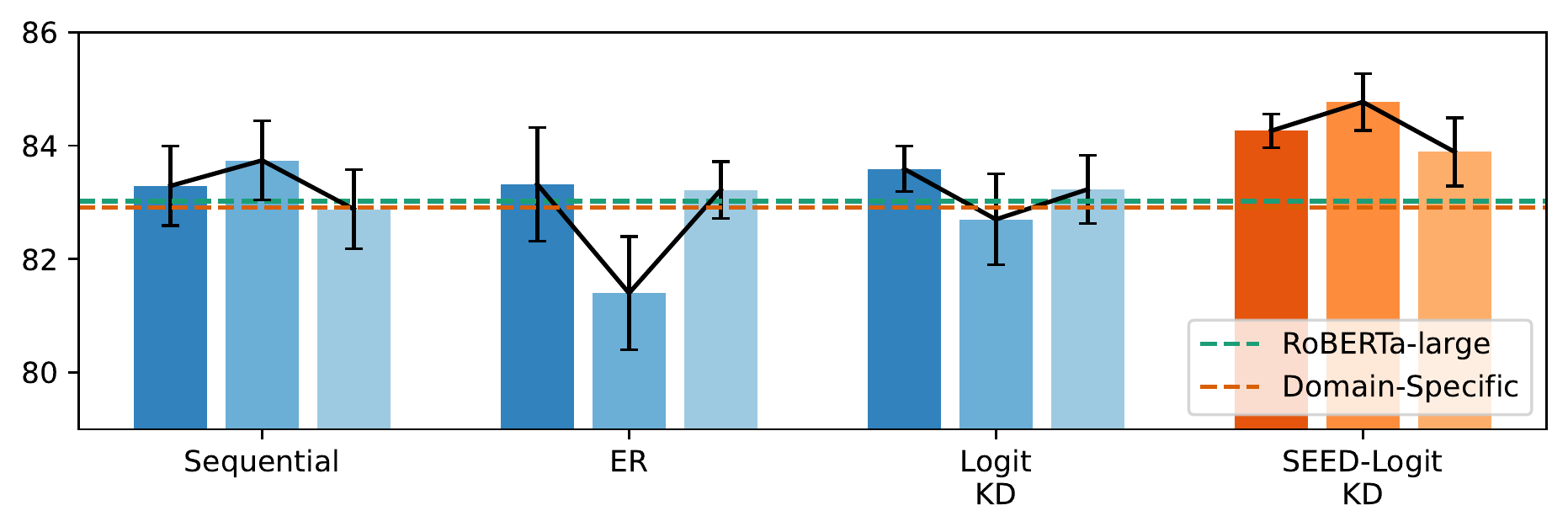}}
    \caption{\small Performance evolution of downstream models with RoBERTa-large as the base model. Models are fine-tuned from checkpoints of lifelong pretrained LMs at different time steps $t$. For Chemprot and RCT-Sample from $D_1$, we use $t \in \{1,2,3,4\}$; while for ACL-ARC and SciERC from $D_2$, $t \in \{2,3,4\}$. Methods achieving the best performance at the end of training ($t=4$) is highlighted.}
    \label{fig:academics_curve_roberta_large}
\end{figure*}

\section{Experiments with RoBERTa-large}
We present additional experiments on RoBERTa-large. Figure~\ref{fig:academics_curve_roberta_large} and Table~\ref{tab:roberta_large} summarizes the results of selected continual learning algorithms and baselines. On Chemprot, RCT-Sample, ACL-ARC and SciERC, either SEED-Logit-KD or Logit-KD achieves best performance with the final pretrained model checkpoint. We notice that sometimes certain continual learning algorithms (ER, Logit-KD) achieves lower F1 at the initial time step  (\textit{e.g.,} $t$=2 in Figure~\ref{fig:academics_curve_roberta_large}(c)). In these cases, we hypothesize continual learning algorithms may hurt model's performance in capturing new knowledge, despite its potential to reduce forgetting.

\begin{table*}[]
\centering
\scalebox{0.60}{
\begin{tabular}{@{}lcccccccccccc@{}}
\toprule
Task          & \multicolumn{3}{c}{$D_1$ - Biomedical} & \multicolumn{3}{c}{$D_2$ - Computer Science} & \multicolumn{3}{c}{$D_3$ - Materials Science} & \multicolumn{3}{c}{$D_4$ - Physics} \\
\cmidrule(r){1-1} \cmidrule(lr){2-4} \cmidrule(lr){5-7} \cmidrule(lr){8-10} \cmidrule(l){11-13} 
Dataset       & Chemprot           & RCT-Sample    &  MLM      & ACL-ARC               & SciERC        & MLM        & MNER                   & Synthesis    & MLM         & Keyphrase         & Hyponym     & MLM     \\ \midrule

Sequential Pretraining         & 82.09$_{\pm 0.5}$          & 79.60$_{\pm 0.5}$      & 1.654    & 72.73$_{\pm 2.9}$             & 81.43$_{\pm 0.8}$    & 1.807         & 83.99$_{\pm 0.3}$            & 92.10$_{\pm 1.0}$    & 1.590          & 67.57$_{\pm 1.0}$         & 74.68$_{\pm 4.4}$    & 1.381    \\

Sequential Pretraining$_{b^\prime=1.2b}$         & 81.68$_{\pm 0.5}$          & 79.80$_{\pm 0.4}$      & 1.656   & 70.57$_{\pm 3.0}$             & 80.89$_{\pm 1.2}$    & 1.793         & 83.65$_{\pm 0.3}$              & 92.16$_{\pm 0.7}$    & 1.578         & \textbf{67.61$_{\pm 1.4}$}         & 75.03$_{\pm 4.1}$    & \textbf{1.379}    \\

ER            & 82.73$_{\pm 0.3}$          & \textbf{79.98$_{\pm 0.3}$}  & 1.737        & 72.50$_{\pm 1.0}$             & 81.64$_{\pm 1.1}$   & 1.857           & 83.99$_{\pm 0.4}$              & \textbf{92.65$_{\pm 0.4}$}     & 1.621        & 66.11$_{\pm 1.1}$         & 72.82$_{\pm 4.3}$   & 1.391      \\

ER$_{k=5}$            &   \textbf{83.00$_{\pm 0.1}$}     &  79.79$_{\pm 0.4}$  &   1.913   &      69.85$_{\pm 2.6}$    &  \textbf{82.30$_{\pm 1.2}$}  &   2.049        &        \textbf{84.03$_{\pm 0.2}$}      &  91.60$_{\pm 0.6}$    &  1.721  &   65.55$_{\pm 0.4}$    &  75.64$_{\pm 3.2}$  & 1.418      \\


Logit-KD-Sparse      &   82.80$_{\pm 0.4}$      &  79.80$_{\pm 0.5}$  &    1.476             &  73.31$_{\pm 2.0}$   & 81.19$_{\pm 0.8}$ &     1.744      &  83.84$_{\pm 0.4}$     &  92.29$_{\pm 0.7}$   &   \textbf{1.472}    &  66.65$_{\pm 0.7}$  & \textbf{77.27$_{\pm 7.1}$} &  1.385 \\
SEED-KD-Sparse      &  82.51$_{\pm 0.4}$         & 79.52$_{\pm 0.5}$  &  \textbf{1.474}       & \textbf{73.70$_{\pm 3.4}$}             & 81.92$_{\pm 0.8}$      & \textbf{1.741}    & 83.96$_{\pm 0.3}$              & 92.20$_{\pm 1.0}$      & 1.480     & 64.75$_{\pm 1.1}$         & 71.29$_{\pm 3.6}$   & 1.381  \\
\bottomrule

\end{tabular}
}
\caption{Performance of distillation algorithms in the setup of controlled computational costs.}
\label{tab:controlled_results}
\vspace{-0.2cm}
\end{table*}

\section{Experiments with BERT on Tweet Stream After 2019}

\begin{table}[]
\centering
\scalebox{0.66}{
\begin{tabular}{@{}lcccc@{}}
\toprule
Task          & 2019-1                 & 2019-2                & 2020-1                 & 2020-2                \\ \midrule
\multicolumn{5}{c}{Hashtag Prediction}                                                      \\ \midrule
BERT-base  & 46.38$_{\pm 0.4}$            & 48.05$_{\pm 0.8}$            & 41.67$_{\pm 1.0}$            & 69.00$_{\pm 0.5}$            \\
Sequential PT        & 50.46$_{\pm 0.1}$            & 52.70$_{\pm 0.7}$            &46.49$_{\pm 1.0}$            &71.63$_{\pm 0.7}$            \\
ER            &49.90$_{\pm 0.4}$            &52.33$_{\pm 0.6}$            &46.84$_{\pm 0.3}$            & 71.67$_{\pm 0.4}$            \\
Logit-KD    &     50.19$_{\pm 0.9}$          &      \textbf{53.70$_{\pm 0.4}$}     &     \textbf{47.64$_{\pm 0.4}$}    & \textbf{72.44$_{\pm 0.5}$}  \\
SEED-Logit-KD    &    \textbf{50.79$_{\pm 0.8}$}          &      52.84$_{\pm 0.5}$     &     46.04$_{\pm 0.4}$    &  72.24$_{\pm 0.6}$  \\
\bottomrule
\end{tabular}
}
\caption{\small Hashtag prediction performance of continually pretrained BERT models over tweets after 2019.}
\label{tab:bert_hashtag}
\end{table}

In this section, we present an additional set of experiments on BERT-base~\cite{Devlin2019BERTPO} model, which is originally pretrained with Wikipedia articles before 2019, with Tweets only after 2019. The training corpora $D_{1..4}$ consist of tweets from the first half of 2019, the second half of 2019, the first half of 2020, and the second half of 2020 respectively. We accordingly construct hashtag prediction and cross-year hashtag prediction datasets. The performance of downstream tasks fine-tuned from the final pretrained model is presented in Table~\ref{tab:bert_hashtag}. We see Sequential PT clearly outperforms BERT-base which is not continually pretrained, and that Logit-KD generally improves hashtag prediction performance compared to Sequential PT except on the first half of 2019. We hypothesize the small temporal gap between $D_{1..4}$ makes improvements less significant than our main experiment setup. We present temporal generalization performance in cross-year hashtag prediction tasks in Table~\ref{tab:bert_hashtag_tg}. Similarly, Logit-KD improves over Sequential PT in two out of three cross-year hashtag prediction setups.

\begin{table}[]
\centering
\scalebox{0.66}{
\begin{tabular}{@{}lccc@{}}
\toprule
Task          & 2019-1$\to$2019-2                 &  2019-1$\to$2020-1      & 2019-1$\to$2020-2                            \\ \midrule
\multicolumn{4}{c}{Hashtag Prediction}                                                      \\ \midrule
BERT-base  & 40.19$_{\pm 0.3}$            & 41.00$_{\pm 0.6}$            & 40.85$_{\pm 0.8}$           \\
Sequential PT        & 43.30$_{\pm 0.7}$            & \textbf{48.60$_{\pm 2.1}$}            &44.07$_{\pm 0.8}$                  \\
ER            &42.96$_{\pm 0.9}$            &46.07$_{\pm 1.6}$            &44.26$_{\pm 0.7}$         \\
Logit-KD    &    43.35$_{\pm 1.6}$          &     46.91$_{\pm 0.5}$     &     \textbf{45.03$_{\pm 0.2}$}     \\
SEED-Logit-KD    &       \textbf{43.56$_{\pm0.4}$}        &      45.77$_{\pm 0.7}$            &  43.76$_{\pm 0.5}$  \\
\bottomrule
\end{tabular}
}
\caption{\small Temporal generalization performance of Hashtag prediction models fine-tuned from continually pretrained BERT models over tweets after 2019.}
\label{tab:bert_hashtag_tg}
\end{table}

\section{Analysis of Data Streams}
\label{apdx:analysis_data_streams}

\begin{figure}[]
    \centering
    \subfloat[\scriptsize{Research Paper Stream}]{\includegraphics[width=0.80\linewidth]{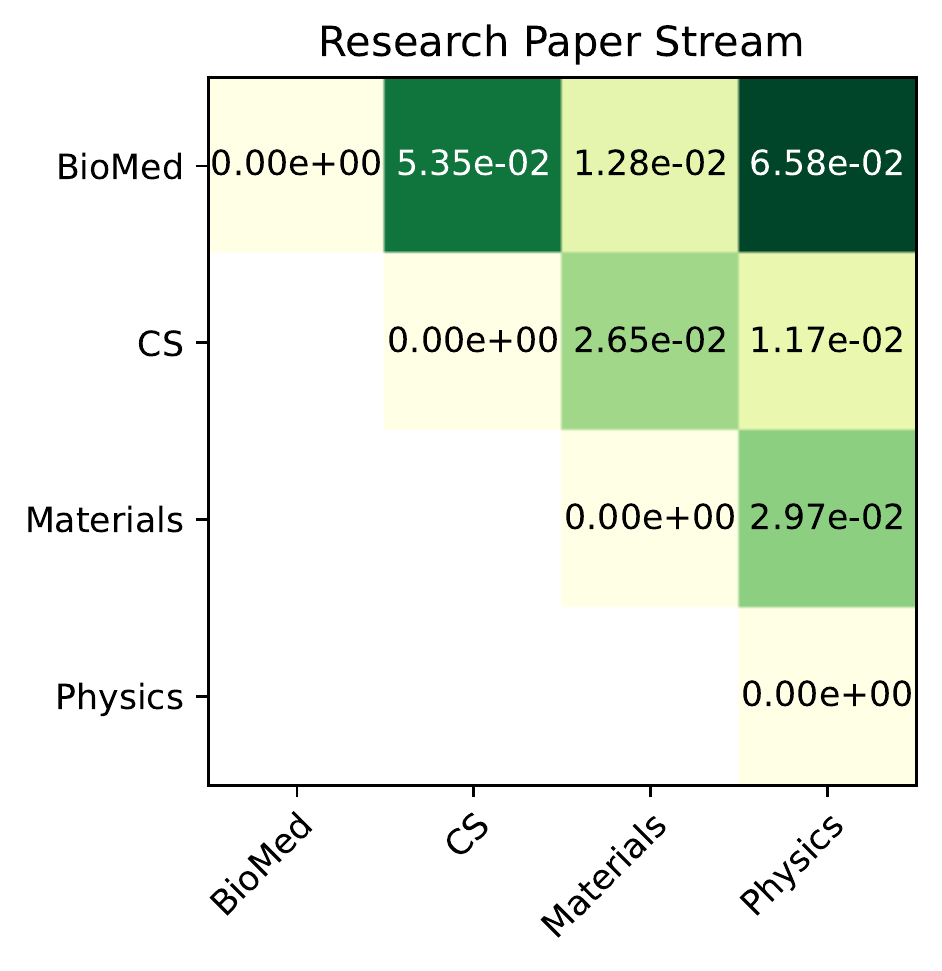}}
    
    \subfloat[\scriptsize{Tweet Stream}]{\includegraphics[width=0.80\linewidth]{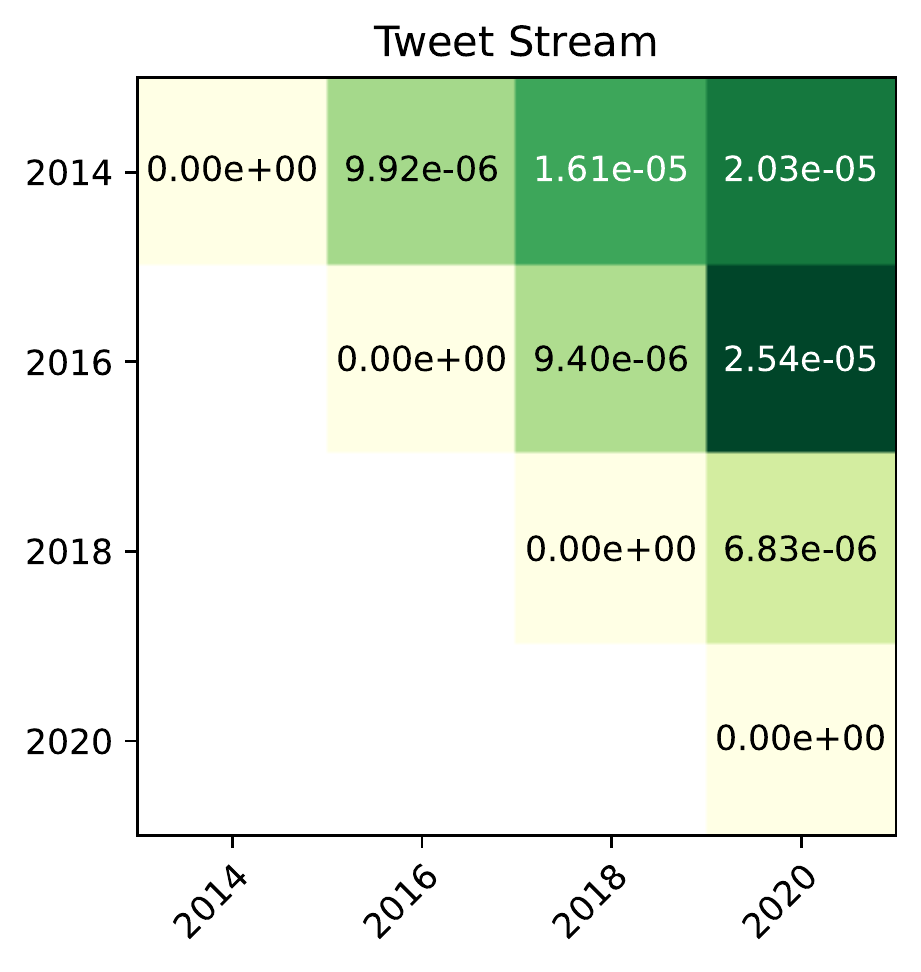}}

    \caption{Cosine distance of vocabulary distributions between each pair of datasets in two data streams.}
    \label{fig:vocab_sim_analysis}
\end{figure}

In this section, we provide further analysis about the created research paper stream and the tweet stream. We measure cosine distances $d_v$ of vocabulary distributions between each pair of different domains $(D_{1..4})$ and summarize the results in Figure~\ref{fig:vocab_sim_analysis}. The results indicate that the Tweet stream has a magnitude smaller vocabulary distribution gap between domains, which is in the scale of $1e^{-5}$, compared to the research paper stream, which is in the scale of $1e^{-2}$. On the Tweet stream, we see the differences of vocabulary distributions align with the temporal gap between domains. On the research paper stream, we find some domains to be more similar than others. For example, Bio-medical ($D_1$) and Material Science domains $(D_3)$ have larger similarity in their vocabulary distributions, which explains general downstream performance increase on $D_1$ after the model is pretrained on $D_3$ (Fig.~\ref{fig:academics_curve} (a,b)).

The differences in vocabulary distribution explain inconsistency in results between two data streams, specifically, whether lifelong pretraining improves downstream model performance on the latest domain, as we mentioned in Sec.~\ref{ssec:temporal_data_stream}.
Other than this, our main findings, such as the effect of distillation-based CL algorithms on reducing forgetting, are consistent over two datasets with such significant differences in their changes of vocabulary distribution. We believe it implies the conclusions in this paper should be reliable in diverse data streams.


\section{Ethic Risks}
We would like to note that, in practice, continually pretrained models over real-world data streams would require identification and removal of biased contents from pretraining corpora, which may affect the prediction of downstream models. As PTLMs are continuously updated, the bias in earlier pretraining may have a profound negative impact. In future works, it is preferable to develop algorithms to ``forget'' certain biased knowledge from language models. We further note that any data released in this paper, especially the tweet stream, should only be used for research purposes.




\end{document}